\documentclass{article} 
\usepackage{collas2022_conference,times}

\usepackage{amsmath,amsfonts,bm}

\def\eqref#1{equation~\ref{#1}}

\def\1{\bm{1}}

\DeclareMathAlphabet{\mathsfit}{\encodingdefault}{\sfdefault}{m}{sl}
\SetMathAlphabet{\mathsfit}{bold}{\encodingdefault}{\sfdefault}{bx}{n}

\usepackage{hyperref}
\hypersetup{
    colorlinks=true,
    linkcolor=red,
    filecolor=magenta,      
    urlcolor=black,
    citecolor=purple,
    pdftitle={},
    pdfpagemode=FullScreen,
    }

\usepackage{url}
\usepackage{amsmath}
\usepackage{mismath}
\usepackage{mathtools}
\usepackage{todonotes}
\usepackage{xcolor}
\usepackage{subcaption}
\usepackage{tikz}
\usetikzlibrary{shadows}
\usepackage{booktabs}
\usepackage{wrapfig}
\usepackage{xcolor}
\usepackage{authblk}

\newcommand{\cM}{{\mathcal{M}}}
\newcommand{\method}[0]{HKS}
\newcommand{\methodfull}[0]{Hierarchical Kickstarting}

\newcommand{\skill}[1]{{\color{blue} \texttt{#1}}}
\newcommand{\task}[1]{{\textsc{#1}}}
\newcommand{\entity}[1]{{\color{red} \texttt{#1}}}

\DeclarePairedDelimiterX{\infdivx}[2]{(}{)}{%
  #1\;\delimsize\|\;#2%
}
\newcommand{\infdiv}{\mathcal{H}\infdivx}

\newcommand*\keystroke[1]{%
  \tikz[baseline=(key.base)]
    \node[%
      draw,
      fill=white,
      minimum width=1.2em,
      drop shadow={shadow xshift=0.25ex,shadow yshift=-0.25ex,fill=black,opacity=0.75},
      rectangle,
      rounded corners=2pt,
      inner sep=2pt,
      line width=0.5pt,
      font=\scriptsize\sffamily
    ](key) {#1\strut}
  ;
}

\title{Hierarchical Kickstarting for Skill Transfer in Reinforcement Learning}

\makeatletter
\renewcommand\AB@affilsepx{ ~ \protect\Affilfont}
\makeatother

\renewcommand*{\Affilfont}{\normalsize\normalfont}

\author[1]{Michael Matthews}
\author[1,2]{Mikayel Samvelyan}
\author[3]{Jack Parker-Holder}
\author[1]{Edward Grefenstette}
\author[1]{Tim Rocktäschel}

\affil[1]{University College London}
\affil[2]{Meta AI}
\affil[3]{University of Oxford}
\affil[ ]{\vspace{0.2cm} \authorcr\url{michael@mtmatthews.com}}

\collasfinalcopy 

\begin{document}

\maketitle
\begin{abstract}
Practising and honing skills forms a fundamental component of how humans learn, yet artificial agents are rarely specifically trained to perform them. Instead, they are usually trained end-to-end, with the hope being that useful skills will be implicitly learned in order to maximise discounted return of some extrinsic reward function. In this paper, we investigate how skills can be incorporated into the training of reinforcement learning (RL) agents in complex environments with large state-action spaces and sparse rewards. To this end, we created SkillHack, a benchmark of tasks and associated skills based on the game of NetHack. We evaluate a number of baselines on this benchmark, as well as our own novel skill-based method Hierarchical Kickstarting (HKS), which is shown to outperform all other evaluated methods. Our experiments show that learning with a prior knowledge of useful skills can significantly improve the performance of agents on complex problems. We ultimately argue that utilising predefined skills provides a useful inductive bias for RL problems, especially those with large state-action spaces and sparse rewards.
\end{abstract}
\section{Introduction}


The acquisition and execution of skills form a fundamental component of how humans learn.  Consider learning to play the game of football. Rather than learning by simply playing successive matches, large parts of training would be commonly devoted to developing specific skills, such as passing, shooting, footwork, and general fitness.  Since humans seem to benefit from explicitly breaking down a complex task into constituent skills, we hypothesise that reinforcement learning (RL) agents can benefit from doing the same.

Most existing methods that incorporate some form of skill-based learning do so by \emph{learning the skills during training}, simultaneously with the policy that makes use of these skills \citep{bacon2016option, frans2017meta, vezhnevets2017feudal}. These two layers of learning often result in instability, although recent approaches have shown success in limiting this \citep{nachum2018dataefficient, levy2019LearningMH}.

In this work, we narrow our focus and consider only the problem of learning with the aid of predefined skills.  Specifically, we assume access to a set of expert policies, one for each skill we have defined.  These experts could be obtained automatically by performing some search for diverse policies \citep{lehman2008novelty, eysenbach2018diversity, dvd}, be hand coded with some heuristic policy, or be trained with RL on a set of skill-specific environments to a level that is deemed sufficient, which is the approach taken in this paper.


The manifestation of this is SkillHack,\footnote{Code available at \url{https://github.com/ucl-dark/skillhack}} a new benchmark for skill-based learning using the MiniHack \citep{samvelyan2021minihack} framework. 
The benchmark consists of 8 unique task environments, each of which has an associated set of skill acquisition environments for mastering the individual skills necessary for completing each task.
The tasks are designed to be of a difficulty such that not utilising the relevant skill environments makes them very hard to solve. 
Completing the suite of tasks requires a broad range of skills in the NetHack environment \citep{kuttler2020nethack} including navigation, combat and manipulation of in-game items.  
The large space of items
, as well as actions for manipulating them, results in a state-action space that is difficult to explore under a random exploration regime.
The range of unique and diverse skills in NetHack, combined with its speed and ease of use, make SkillHack a convenient benchmark for studying policy transfer in RL, as well as problems such as continual learning and unsupervised RL.


\begin{figure}[ht!]
    \centering
    \includegraphics[width=.7\textwidth]{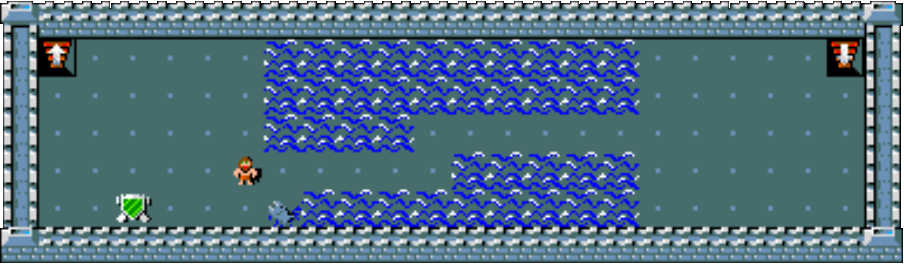}
    \caption{\textsc{Sea Monsters} task.  The agent must reach the staircase on the opposite bank without dying to the monsters in the lake.  To do this it must use of the powerful suit of armour that has spawned on the near bank.  The agent can most easily learn to complete this task by transferring knowledge gained from its associated skill environments.}
    \label{fig:sea_monsters}
\end{figure}%
\begin{figure}[ht!]
\centering
\begin{subfigure}{.32\textwidth}
    \centering
    \includegraphics[width=0.9\textwidth]{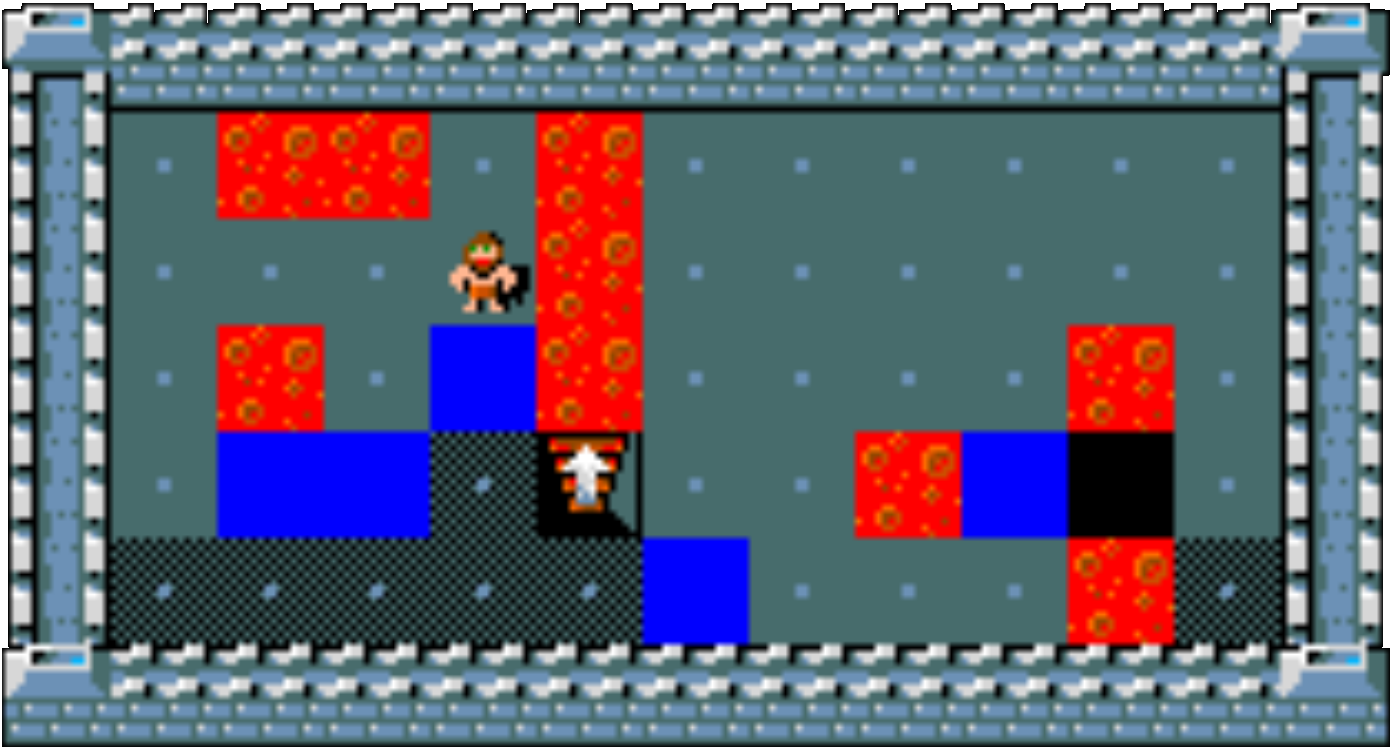}
    \caption{\skill{TakeOff}: The agent must take off the clothes it starts with by pressing \keystroke{Shift}+\keystroke{t}.}
\end{subfigure}%
\begin{subfigure}{.32\textwidth}
    \centering
    \includegraphics[width=0.9\textwidth]{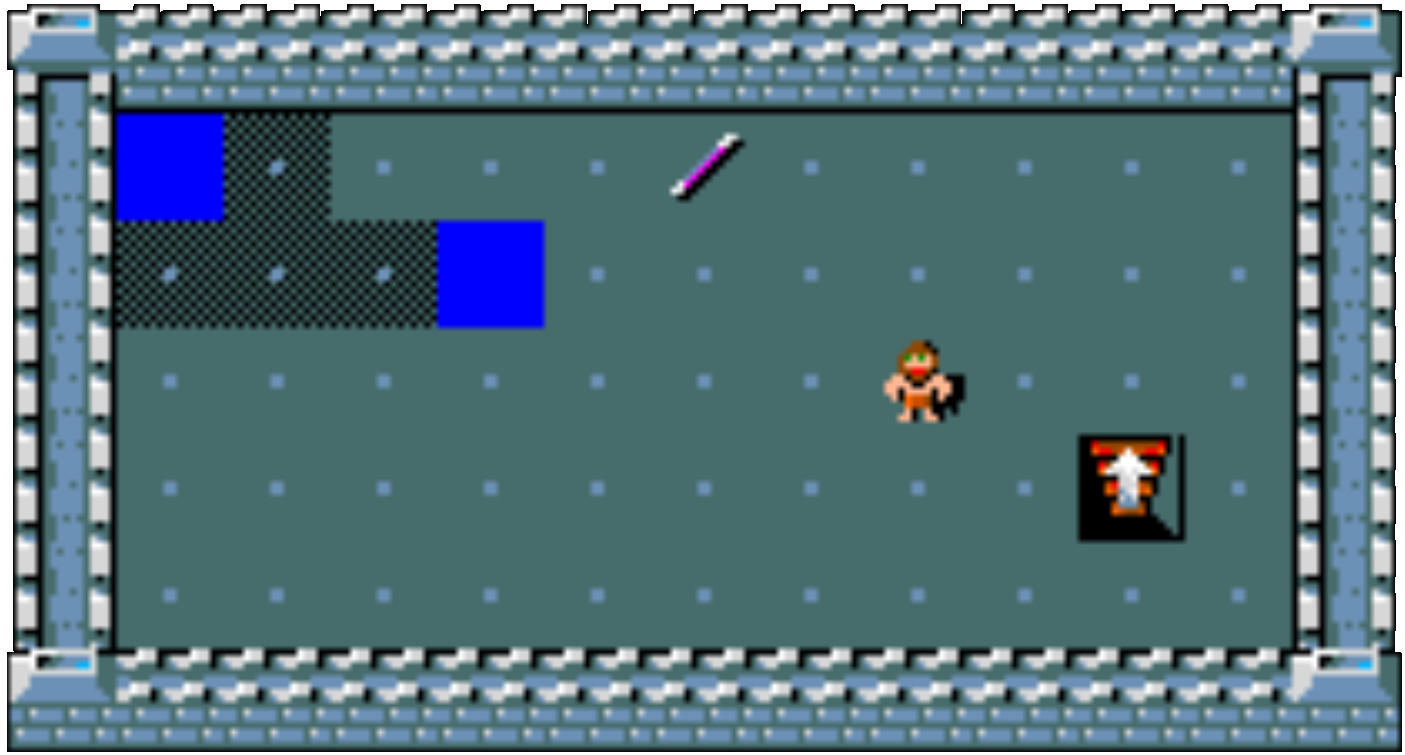}
    \caption{\skill{PickUp}: The agent must navigate to the item and pick it up with \keystroke{,}.}
\end{subfigure}%
\begin{subfigure}{.32\textwidth}
    \centering
    \includegraphics[width=0.9\textwidth]{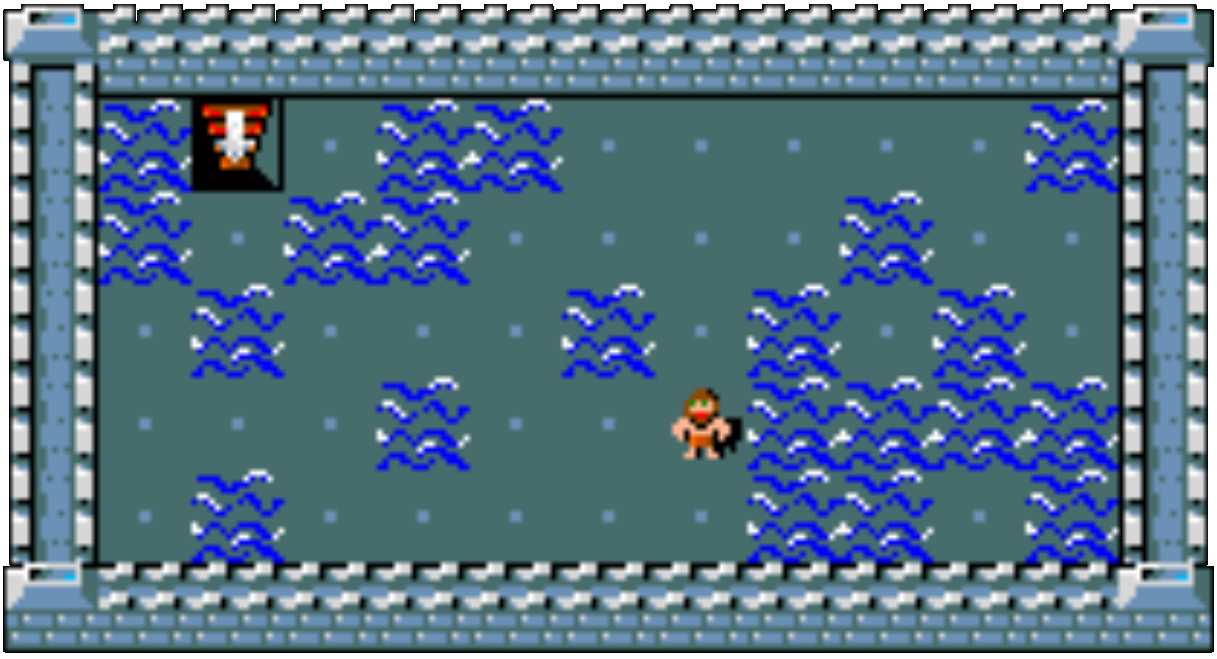}
    \caption{\skill{NavigateWater}: The agent must reach the staircase, receiving a reward penalty every time it tries to walk into the water.}
\end{subfigure}
\caption[short]{Some of the skill environments for the skills associated with the \task{Sea Monsters} task.}
\label{fig:sea_monster_skills}
\end{figure}

To evaluate SkillHack, we propose \methodfull{} (\method), a new policy transfer approach which distills knowledge acquired from expert policies trained on the skill acquisition environments. We refer to these expert policies as \textit{teachers}. 
\method{} learns a policy-over-teachers $\pi_H$, which outputs a weighting for each teacher at every timestep.  These are then used to calculate a weighted average of the action distribution of the teachers.
The agent policy $\pi$ then incurs a loss proportional to the divergence of its own action distribution from this weighted average. 
This incentivises the agent to behave more similarly to highly weighted teachers, potentially accelerating learning by nudging the agent to behave more similarly to a useful policy. 
Simultaneously, the policy-over-teachers is incentivised to pick teachers that match the agent. 
After a burn-in time, these two networks should be in sync, with the policy-over-teachers recommending useful teachers based on the current state of the agent, while the agent behaves similarly to the chosen teachers.
This method can be considered as a more generalised version of Kickstarting \citep{schmitt2018kickstarting}, where we add the network $\pi_H$ so that different skills can be prioritised for knowledge transfer at different points within an episode.

We perform an empirical evaluation of \method{} as well as several baselines across all SkillHack environments. Our results show that learning with a prior knowledge of useful skills can significantly improve performance of agents on complex problems. 
We demonstrate that while \method{} outperforms other baselines on the overall suite of SkillHack tasks, there is still significant room for methods to improve.

In summary, this paper makes the following contributions:
(i) we present SkillHack, a new benchmark for skill transfer in RL,
(ii) we propose \method{}, a new method for policy transfer based on skill specific expert policies,
(iii) we provide an evaluation and discussion of \method{} and several other baseline methods on the SkillHack benchmark. 

\section{Background}

\subsection{Reinforcement Learning}

We use the standard formalism of Markov Decision Process (MDP) defined as a tuple $M = \langle \mathcal{S}, \mathcal{A}, \mathcal{T}, \mathcal{R}, \gamma \rangle$ representing state space $\mathcal{S}$, action space $\mathcal{A}$, transition probability $\mathcal{T}:\mathcal{S} \times \mathcal{A} \xrightarrow{} \mathcal{P}(\mathcal{S})$, reward function $\mathcal{R}: \mathcal{S} \times \mathcal{A} \xrightarrow{} \mathbb{R}$, and discount factor $\gamma$. 
The agent chooses actions according to a stochastic policy $\pi: \mathcal{S} \xrightarrow{} \mathcal{P}(\mathcal{S)}$ in order to maximise expected episodic reward, defined as $\mathbb{E}\left[\sum_{t=0}^{T}\gamma^{t}r_{t}\right]$ where $T$ is the final timestep and $r_t$ is reward at timestep $t$. 
The value function of policy $\pi$ is defined as $v_\pi(s) = \mathbb{E}_\pi \Big[ \sum_{t=0}^{T}\gamma^{t} r_{t} \vert s_t = s \Big]$.
Model-free approaches to RL aim to learn the optimal policy without explicitly learning the dynamics of the environment.
Actor-critic algorithms \citep{konda2000actor} make use of both a policy (actor) for control and an estimated value function  (critic) for updating the policy of the agent.
IMPALA \citep{espeholt2018impala} scales up actor-critic to work with multiple actors, each interacting with their own environment.  These actors then periodically send trajectories of experience to a centralised learner, which then updates the policy and communicates back to the actors.  Since the policy of the actors will lag behind that of the centralised learner, the update uses V-trace, an off-policy corrected target value.

\subsection{Policy Transfer}

Training RL agents from scratch can often be prohibitively expensive and time-consuming, while making use of existing knowledge can improve sample efficiency and training speed. \textit{Policy transfer} approaches aim to assist the learning process of the agent by utilizing pretrained policies on related tasks.
We refer to the policy designed for the target task $M_T$ as the \textit{student}, which can leverage knowledge from \textit{teacher} policies $\pi_{E_1}, \pi_{E_2}, \dots, \pi_{E_K}$ trained on a set of source domains $\cM_1, \cM_2, \dots, \cM_K$, respectively. 




%

\paragraph{The Options Framework}

One way to perform policy transfer is to use options \citep{sutton1999smdp}, a Hierarchical Reinforcement Learning (HRL) framework that introduces temporally-extended actions that can span multiple timesteps. An option $\omega$ is a 3-tuple $\langle \mathcal{I}, \pi, \beta \rangle$ representing the states the option can be initiated from, the option policy and the option termination function.  In a given state $s$, a policy-over-options selects an option $\pi_\omega$ for which $s \in \mathcal{I}_\omega$. Control over actions is then ceded to this option for a number of timesteps.  Once the termination condition $\beta_\omega$ is stochastically met, the option finishes and cedes control back to the policy-over-options, which then selects a new option to execute.



\paragraph{Kickstarting RL}

Kickstarting \citep{schmitt2018kickstarting} is a policy transfer method that uses one or more expert teachers to guide a student policy. Multi-teacher Kickstarting works by adding an auxiliary loss to the student during training, defined as $l_\text{kick} = \infdiv*{\sum_k{\lambda_k \pi_k(\cdot | s_t)}}{\pi (\cdot | s_t)}$, where $\infdiv*{\cdot}{\cdot}$ is cross-entropy, $\pi_k$ are the teacher policies, $\pi$ is the student policy and $\lambda_k$ are the per-teacher Kickstarting coefficients.  This means that the agent not only optimises for extrinsic reward, but is also incentivised to behave similarly to the teachers.  The $\lambda_k$ can be varied over training (although remain constant within an episode) either with Population-Based Training \citep{jaderberg2017population} or with a manual schedule. 



\subsection{NetHack}

The NetHack Learning Environment \citep{kuttler2020nethack} is an RL environment based off the classic game of \textit{NetHack}. NetHack is a dungeon crawler game, notorious for its difficulty, where the player must work their way through dozens of procedurally generated levels, with hundreds of unique enemies and objects.  The game is turn based and stochastic, with a large action space.
%
%
%
MiniHack \citep{samvelyan2021minihack} is an extension of the NetHack Learning environment that allows for customisation of levels, rewards and termination conditions, while retaining access to the full set of entities and environment dynamics from NetHack.  

\begin{table*}[ht]
    \centering
    \caption{Summary of SkillHack tasks.  Different styles are used to differentiate \task{Tasks}, \skill{Skills} and \entity{Items/Entities}.}
    \begin{tabular}{p{0.2\linewidth} p{0.73\linewidth}} 
     \toprule
     \textbf{Task} & \textbf{Description}\\
     \midrule
    \task{Battle} & \hspace{0pt}\skill{PickUp} a randomly placed \entity{Sword}, \skill{Wield} the \entity{Sword} and finally \skill{Fight} and kill a \entity{Monster}. \\
     \midrule
     \task{Prepare for Battle} & \hspace{0pt}\skill{PickUp} a ration of \entity{Food} and a piece of \entity{Armour}.  \skill{Wear} the \entity{Armour} and \skill{Eat} the \entity{Food}.\\
     \midrule
     \task{Target Practice} & Either a set of \entity{Daggers} or a \entity{Wand of Death} will spawn on the floor.  \skill{PickUp} the item and either \skill{ZapWandOfDeath} or \skill{Throw} the \entity{Daggers} in order to kill the \entity{Minotaur}.\\
     \midrule
     
     \task{Medusa} & \hspace{0pt}\skill{PickUp} and then \skill{PutOn} the \entity{Towel} that is placed in your starting room, thus blindfolding yourself.  Open the door to the second room where \entity{Medusa} is caged, but can still instantly kill you with her deadly gaze should you enter with your vision intact. You must \skill{NavigateBlind} in this second room in order to find the stairway out.\\
     \midrule
     \task{Sea Monsters} & \hspace{0pt}\skill{TakeOff} your starting \entity{Armour} and then \skill{PickUp} and \skill{Wear} the strong suit of \entity{Armour} that spawns on the near bank of the lake.  Whilst wearing this \entity{Armour} you can \skill{NavigateWater} across the bridge without dying to the \entity{Piranhas}.  Make it to the other side and reach the staircase to complete the task.\\
     \midrule
     \task{Frozen Lava Cross} & \hspace{0pt}Either a \entity{Wand of Cold} or a \entity{Frost Horn} will spawn on the near side of a river of lava.  \skill{PickUp} the item and then create a bridge across the lava with either \skill{ZapWandOfCold} or by \skill{ApplyFrostHorn}.  Finally, \skill{NavigateLava} across your newly made bridge to reach the staircase on the other side.\\
     \midrule
     \task{Identify Mimic} & \hspace{0pt}\entity{Mimics} are monsters that can camouflage themselves.  This task generates 3 \entity{Statues}, 2 of which are real and one of which is actually a \entity{Mimic} that you must safely identify.  To do this you can \skill{PickUp} a stack of \entity{Daggers} and \skill{Throw} them at the \entity{Statues}.  Hitting the \entity{Mimic} will reveal it and complete the task.  Alternatively you can \skill{NavigateLavaToAmulet} by following the bridge across the lava lake and then \skill{PickUp} and \skill{PutOn} the \entity{Amulet of ESP} which will reveal the \entity{Mimic}.  Revealing the \entity{Mimic} in an unsafe way, for instance by walking into it, will cause you to fail the task.\\
     \midrule
     \task{A Locked Door} & \hspace{0pt}\skill{PickUp} the \entity{Skeleton Key} and use it to unlock the \entity{Door}, before proceeding to \skill{NavigateLava} to the exit. \\
     \bottomrule
    \end{tabular}
    \label{tab:tasks}
\end{table*}

\section{SkillHack: A benchmark for skill transfer}

To serve as a benchmark for skill transfer we developed \textit{SkillHack}: a set of procedurally generated MiniHack environments containing 8 task environments and 16 skill acquisition environments.  The task environments are designed to be tricky to solve with vanilla RL, due to the large state-action space and sparse rewards.  For this reason, each task environment has an associated set of relevant skill environments.  By transferring knowledge gained from solving these skill environments, the task environment will become easier to solve.

Each task environment is set up such that the agent receives a $+1$ reward for successfully completing the task and a $-1$ reward for failing the task and/or dying.  The skill environments are set up with bespoke rewards\footnote{Most of the skill environments simply have +1 reward for completing the skill.  More complicated skills, for instance \skill{Unlock}, require an exact sequence of key presses.  These environments have additional rewards for getting part way through the sequence.} to encourage an RL agent to learn the skill quickly and in a generalised manner.

A motivating example is the \task{Sea Monsters} task (Figure \ref{fig:sea_monsters}).  In this task, the agent is faced with a narrow, procedurally generated bridge over a monster infested lake, with the goal to reach the staircase on the opposite bank. If the agent simply walks over the bridge, it will be killed by one of the monsters before it reaches the other side. However, on the near side of the lake a powerful piece of armour will be randomly generated and placed on the floor.  The agent must \skill{Take Off} the clothes it starts off wearing, \skill{Pick Up} and then \skill{Wear} the armour, before finally \skill{Navigating Water} to reach the staircase on the other side of the lake. 
 
The four skills associated with the task each have their own respective environments (Figure \ref{fig:sea_monster_skills}), in which skill-specific teacher agents can learn expert policies.  The environments are all procedurally generated, with distractions in the form of random terrain and entities, in order to prevent overfitting \citep{cobbe2020leveraging}.  As well as providing reward and termination when the skill is successfully performed, a constant negative reward is applied every timestep in order to encourage the agent to perform the skill as fast as possible.

It should be noted that these task and skill acquisition environments only form a fraction of the possible  behaviours in NetHack. However, the primary aim of SkillHack is not necessarily to directly lead to a more competent NetHack agent, but to facilitate research into skill transfer.


The full list of tasks is summarised in Table \ref{tab:tasks}.  More information on the skill acquisition and task environments can be found in appendices \ref{app:skill_listing} and \ref{app:task_listing} respectively.
\section{Hierarchical Kickstarting}

In this section, we propose \methodfull{} (\method{}), a new approach for policy transfer that combines the strengths of two other policy transfer approaches: Kickstarting and the Options Framework.

To perform policy transfer from multiple teachers, \method{} utilises a hierarchical policy network $\pi_H$ which at every timestep weights the relevance of each teacher $\pi_k$ based on the current state. 
Rather than ceding control to one of the teacher policies $\pi_k$ as in the Options Framework, this weighting is used to calculate a weighted average of the teacher action distributions.  We then formulate an additional loss, proportional to the cross-entropy of this weighted average with the agent action distribution.
The policy-over-teachers $\pi_H$ therefore does not directly affect the action selection but only the loss incurred by the agent. 
As a result, the student $\pi$ is optimised to behave similarly to the average of the teachers $\pi_k$ as weighted by $\pi_H$, as well as maximising external rewards. 
%
%
%
Conversely, $\pi_H$ is optimised to choose teachers that match the behaviour of $\pi$, forming a cyclic relationship between the two levels of policies. 

The additional auxiliary loss of the student policy in \method{} is computed as follows:
\begin{align}
    l_\text{\method{}} = \lambda \infdiv*{\sum_k{\pi_H(k | s_t) \pi_k(\cdot | s_t)}}{\pi (\cdot | s_t)},
\end{align}
here the outputs of $\pi_H$ are passed through a softmax function.

\method{} can be seen as a combination of Kickstarting and the Options Framework. It generalises Kickstarting by using a hierarchical policy-over-teachers which adjusts how much to kickstart from each teacher at each timestep, rather than distilling a fixed amount from each teacher.  Since skills often have to be performed sequentially to complete a task, it is common for only a single skill to be immediately relevant at a given point in time, making this a useful ability.

Furthermore, \method{} also addresses the weakness of the Options Framework. Pretrained options are effective when the environment dynamics of the target domain matches those used during the training of the option. However, options can be highly unstable when executed directly on a target task which is meaningfully different from the environment encountered during training. 

\section{Experimental Setup}

All experiments are run with the IMPALA \citep{espeholt2018impala} framework, using the open-source TorchBeast implementation \citep{torchbeast2019}.  The agent architecture is an adapted version of the one used in \citet{kuttler2020nethack} and \citet{samvelyan2021minihack}.  

The inputs to the network include the full matrix of \textit{glyphs} that serve as the main screen of the game, along with an encoding of the most recently received in-game message (``\textit{You hit the orc!}"), the players inventory and a set of relevant statistics (health, armour level, etc.).  The network uses CNNs for the spatial inputs, as well as LSTM units to incorporate memory.  The action space consists of 32 actions (16 movement actions and 16 command actions).  For the full action space see Appendix \ref{app:nn} and for a full listing of hyperparameters see Appendix \ref{app:hyp}.  For more detail on the observation space and network architecture see \citet{kuttler2020nethack} and \citet{samvelyan2021minihack}.

An expert teacher policy is produced via vanilla RL for each of the 16 skill environments.  The results of training the experts can be seen in Appendix \ref{app:further_results}.  We train vanilla RL, Random Network Distillation (RND) \citep{burda2018rnd}, the Options Framework, Kickstarting and \method{} on all 8 tasks for $2 \times 10^8$ timesteps, repeated over at least $3$ random seeds.  For the 3 skill-based methods, only teachers relevant to the target task are used. 

\subsection{Baselines} \label{sec:baselines}

\paragraph{Options Framework} We use the pretrained experts as options with frozen weights, while training a policy-over-options network to direct them.  The option networks have their LSTM units removed (although these are kept in the policy-over-options).
We consider a degenerate form of the Options Framework where options can be initialised from any state and always terminate in one step $\mathcal{I}_\omega = \mathcal{S}$, $\beta(\cdot)_\omega=1$.

\paragraph{Kickstarting} 
For the Kickstarting agent, we use the experts as teachers and similarly remove the LSTMs.  We keep all $\lambda_k$ equal and therefore refer to them all as $\lambda=\lambda_k$.  The schedule for $\lambda$ was manually set to start at $\lambda=10$ and linearly decay to $\lambda=1$ after $10^7$ timesteps.  Retaining a small Kickstarting coefficient was found to increase the stability of the method, whereas if $\lambda$ was allowed to decay to 0 there would be occasional massive drops in performance (Appendix \ref{app:ks_sched}).

\paragraph{Hierarchical Kickstarting}
The LSTM removal and $\lambda$ schedule are the same as for Kickstarting.  The policy-over-teachers was implemented as an extra head on the student network in the form of an additional single fully connected layer with softmax activations.  Preliminary testing of the method found that it would often quickly get stuck in local minima where the policy-over-teachers would always weight the same teacher very highly, resulting in poor performance.  To mitigate this, an additional loss was added proportional to the negative entropy of the policy-over-teachers $l_E = -\kappa\mathcal{H} (\pi_H)$.  This encourages the policy-over-teachers to give a more uniform distribution, preventing it from falling into the described local minima.  The coefficient $\kappa$ was set with a manual schedule, starting at $\kappa=20$ and linearly decaying to $\kappa=0$ after $2\times10^7$ timesteps.

\textbf{Vanilla RL} and \textbf{RND} did not require any additional changes.

\section{Results and Discussion} \label{sec:results}
\subsection{Results}
\begin{wrapfigure}{r}{0.5\textwidth}
\vspace{-0.5cm}
  \begin{center}
    \includegraphics[width=0.45\textwidth]{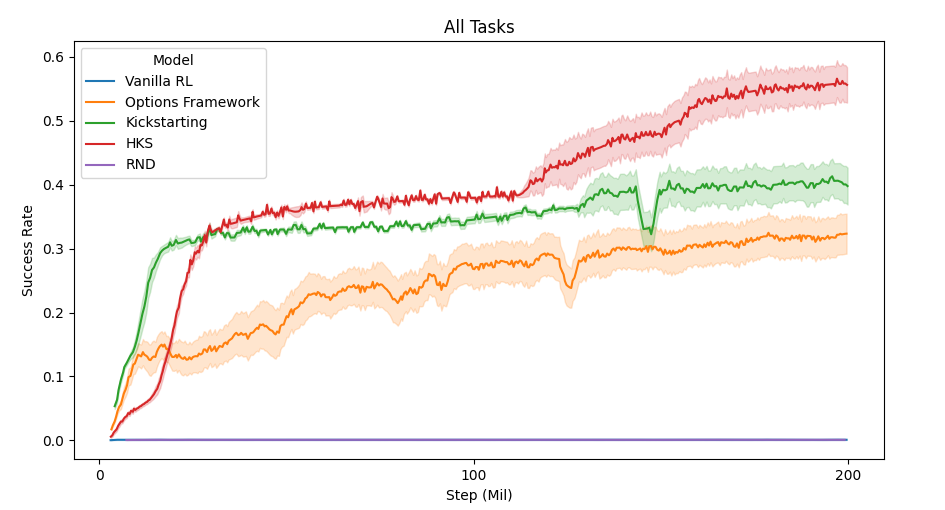}
  \end{center}
    \caption{Mean success rate of baselines across all SkillHack tasks.  The shaded area denotes 1 standard error.}
    \label{fig:results_summary}
    \vspace{-0.2cm}
\end{wrapfigure}


Figure \ref{fig:results_summary} plots the success rate of the 4 methods averaged over all 8 of the SkillHack tasks.  The complete lack of progress by vanilla RL, in contrast to the progress by the 3 skill-based methods, shows that the SkillHack tasks are suitably designed to be a useful benchmark for policy transfer.  HKS performs best in terms of final performance, achieving $56\%$ success rate, as opposed to $41\%$ and $32\%$ for Kickstarting and the Options Framework, respectively.  It does however take marginally longer than the other methods to converge.  HKS overtakes the Options Framework and Kickstarting in success rate at around $2 \times 10^7$ and $3 \times 10^7$ timesteps respectively.  For a fair comparison against HKS, the Kickstarting score was increased by around $2\%$ for reasons discussed in Appendix \ref{app:ks_sched}.


Figure \ref{fig:results} splits out the results for each of the SkillHack tasks, painting a more nuanced picture of the results.  There are some environments where only a subset of the skill-based methods managed to achieve any significant performance, namely \task{A Locked Door} (HKS), \task{Target Practice} (Options Framework) and \task{Medusa} (Kickstarting \& HKS).  In terms of the final success rate, HKS always at least equals Kickstarting, although notably loses to the Options Framework on \task{Target Practice}.  
Interestingly, the only two tasks where the Options Framework outperforms Kickstarting (\task{Target Practice} \& \task{Frozen Lava Cross}) are the only two tasks that cannot always be solved using the same method for each episode.  Both tasks need to be solved differently depending on a random object spawned at the start of the episode (See Table \ref{tab:tasks}).
The \task{Sea Monsters} task was the only one to remain unsolved by all 4 methods, making it a good candidate for future research into skill-based methods.  

\begin{figure}
    \centering
    \includegraphics[width=1\textwidth]{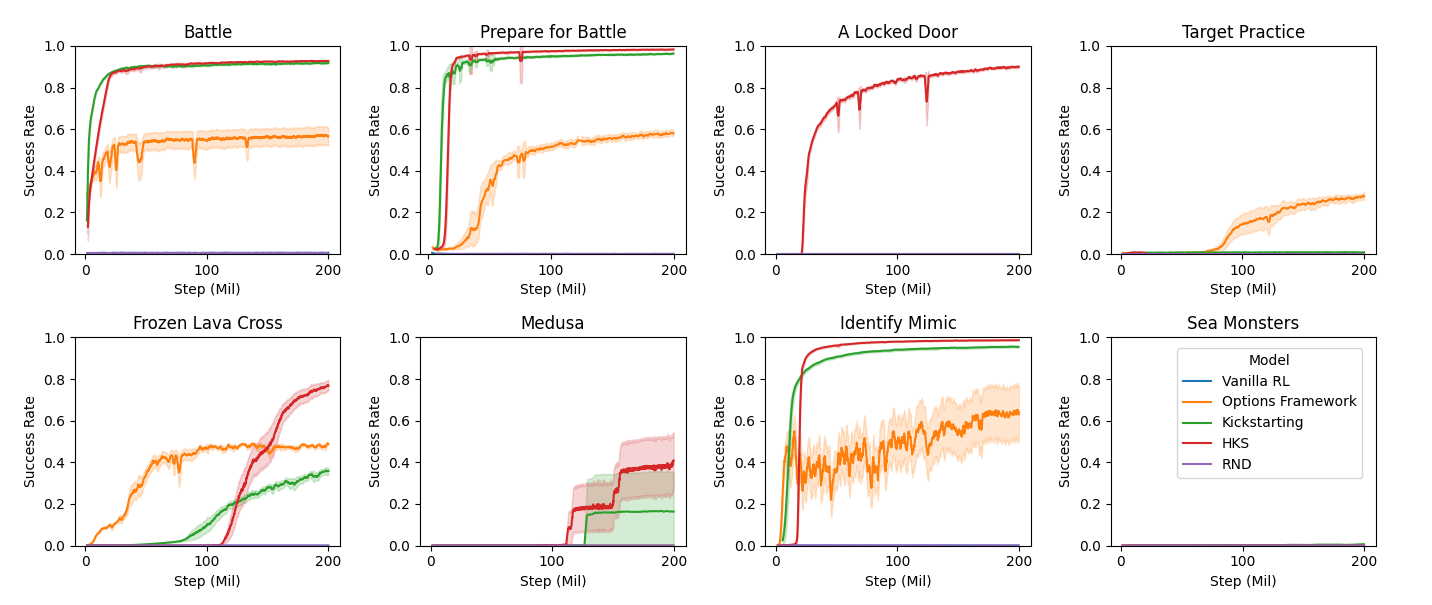}
    \caption{Success rate for all methods on the 8 SkillHack tasks, averaged over the repeats of each experiment.  The shaded area denotes 1 standard error.}
    \label{fig:results}
\end{figure}

\subsection{Mismatched Skills}

\begin{figure}
    \centering
    \includegraphics[width=1\textwidth]{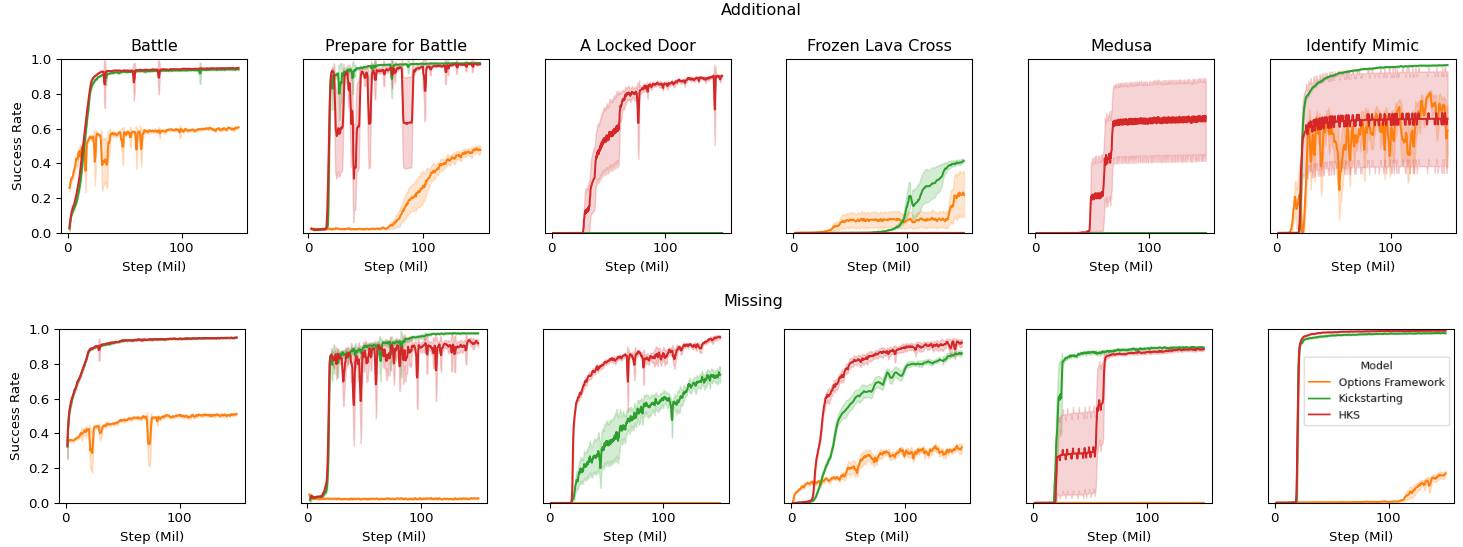}
    \caption{Success rate of the 3 skill based methods on 6 of the SkillHack tasks, when either a useless skill has been added (first row) or a useful skill has been removed (second row) from the skill set.}
    \label{fig:mismatched}
\end{figure}

So far we have only considered the case where the provided skills exactly match those of the task.  In practice, it would likely be a common situation that the skills do not perfectly correspond to the task at hand.  To investigate this, we look at how the skill based algorithms perform in two situations: adding a useless skill and removing a useful skill.  The results are shown in Figure \ref{fig:mismatched}.

The Options Framework performs only marginally worse when given an additional skill, compared to a perfect skill set, which makes sense as it has to navigate a slightly larger action space.  It performs significantly worse when a skill is missing, since this fundamentally limits its capability, as it cannot learn new behaviours.  Surprisingly, both Kickstarting and HKS actually perform better when a skill is missing compared to when an useless skill is added (note that we may expect the result to be different for Kickstarting if PBT was implemented).  We hypothesise that the reason for this is that the remaining skills already give a strong enough prior on exploration, while the existence of an additional skill effectively dilutes this prior.  Interestingly, this implies that more focus should be put on making sure all skills in the skill set are relevant, rather than developing a skill set that covers all possibly useful behaviour.  In practical terms, this may mean developing a small core set of skills that exhibit known essential behaviour.

\subsection{Qualitative Analysis of HKS} \label{sec:qual_eval_hks}

\begin{figure}
    \centering
    \includegraphics[width=0.8\textwidth]{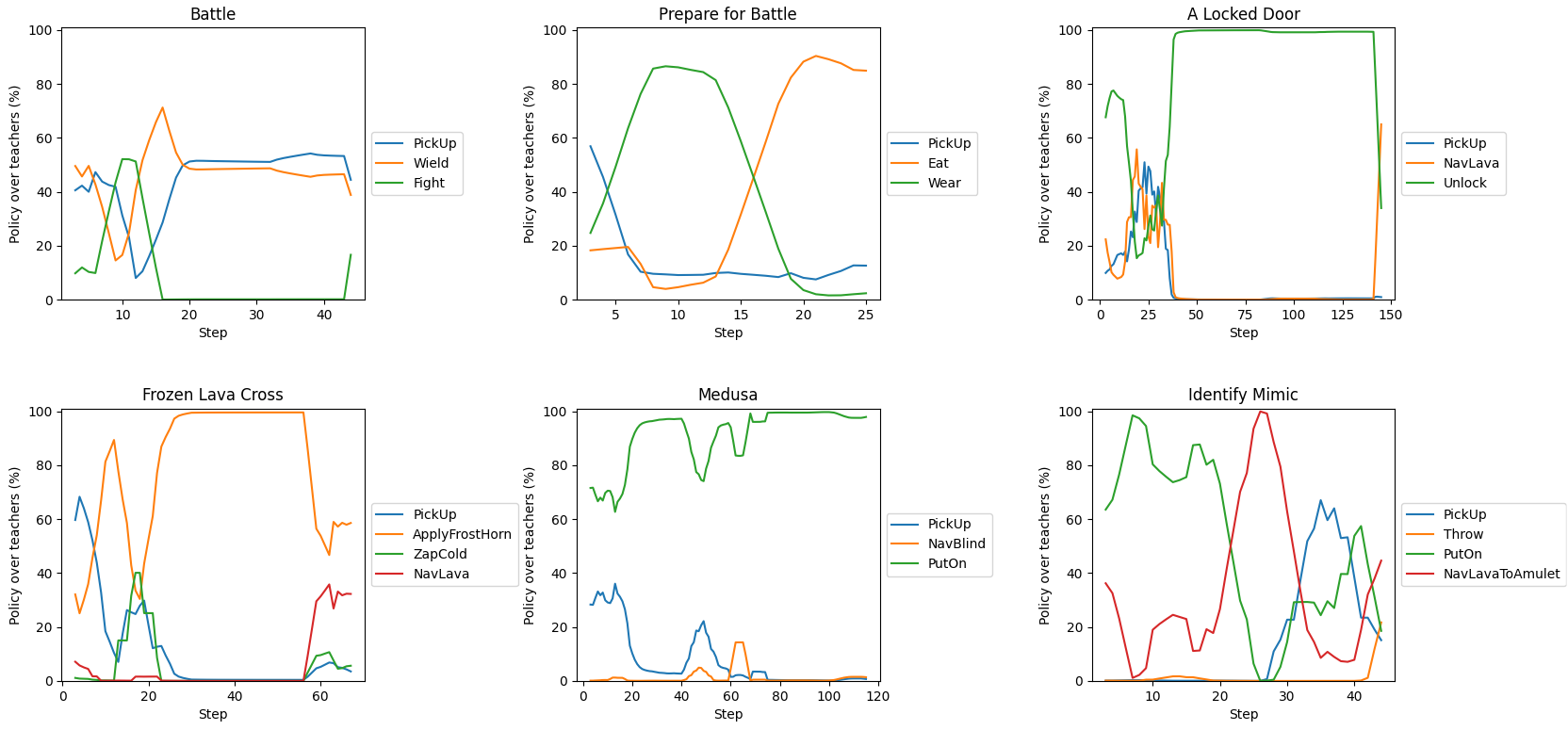}
    \caption{Distribution of policy-over-teachers for converged HKS policies over the span of a single episode.  Results are each taken from a single successful run, smoothed with a rolling average of 6 timesteps.  Note that the sampled \task{Frozen Lava Cross} had a \entity{Frost Horn} spawn.  Only the 6 environments in which HKS converged on are shown.}
    \label{fig:pot_results}
\end{figure}

In order to further scrutinise HKS, we can observe how the distributions of the policy-over-teachers change during an episode for the converged policies (Figure \ref{fig:pot_results}).  Note again that the policy-over-teachers does not affect the agents choice of action, but due to the system of losses implemented in HKS, it can be used to identify which skills the agent is prioritising at each timestep.

\textbf{Sequential Skill Selection} All the given tasks can be solved by executing a set of skills in sequence, so we would expect to see the policy-over-teachers reflect this by prioritising each skill as it becomes relevant.  This behaviour is indeed observed and is most clearly seen in \task{Prepare For Battle}, \task{A Locked Door} and \task{Frozen Lava Cross}, while being somewhat present in \task{Battle}.  The more interesting cases are when this behaviour does not occur.

\textbf{\task{Identify Mimic}: A Missing Skill}  While the policy-over-teachers for \task{Identify Mimic} seems to display the expected behaviour by successively prioritising different skills, on closer inspection the chosen skills do not match the task description.  Specifically, the agent prioritises the \skill{PutOn} skill for the first 20 timesteps, long before it reaches the \entity{Amulet of ESP} and has any need to \skill{PutOn} anything.  These first 20 timesteps are the time when the agent is navigating past the statues in order to make it to the lava bridge (See Figure \ref{fig:app_identify_mimic}).  It would appear that the agent is effectively missing a skill \skill{NavigatePastStatues} and in order to minimise the HKS loss, the policy-over-teachers \textit{prioritises the existing skill that is most similar to the missing one}.  

\textbf{\task{Medusa}: A Useless Skill}  \task{Medusa} also exhibits unexpected behaviour.  In the final timesteps, at the time when the agent should be navigating while blindfolded towards the staircase, we see that the policy-over-teachers overwhelmingly prioritises \skill{PutOn} rather than \skill{NavigateBlind}.  Furthermore, watching the trained agent showed it solved the environment in the expected way and did not find some unexpected alternate solution.  Upon further investigation, we found that \skill{NavigateBlind} does not properly correspond to the \task{Medusa} task.  Specifically, the blindness applied in \skill{NavigateBlind} uses a \entity{Potion of Blindness}, which has different mechanics to blindfolding via \skill{PuttingOn} a \entity{Towel},\footnote{\url{https://nethackwiki.com/wiki/Blindness}} corresponding to a different input observation.  This domain shift made \skill{NavigateBlind} a useless skill\footnote{A fixed version of this skill is available in the GitHub repository.  The authors decided to retain the broken version in this paper as an illustrative point.} and HKS correspondingly learned to disregard it by again prioritising the skill during this time period that most closely matched the theoretical correct \skill{NavigateBlind} skill.

\subsection{Analysis of HKS Loss}

\begin{figure}
    \centering
    \includegraphics[width=1\textwidth]{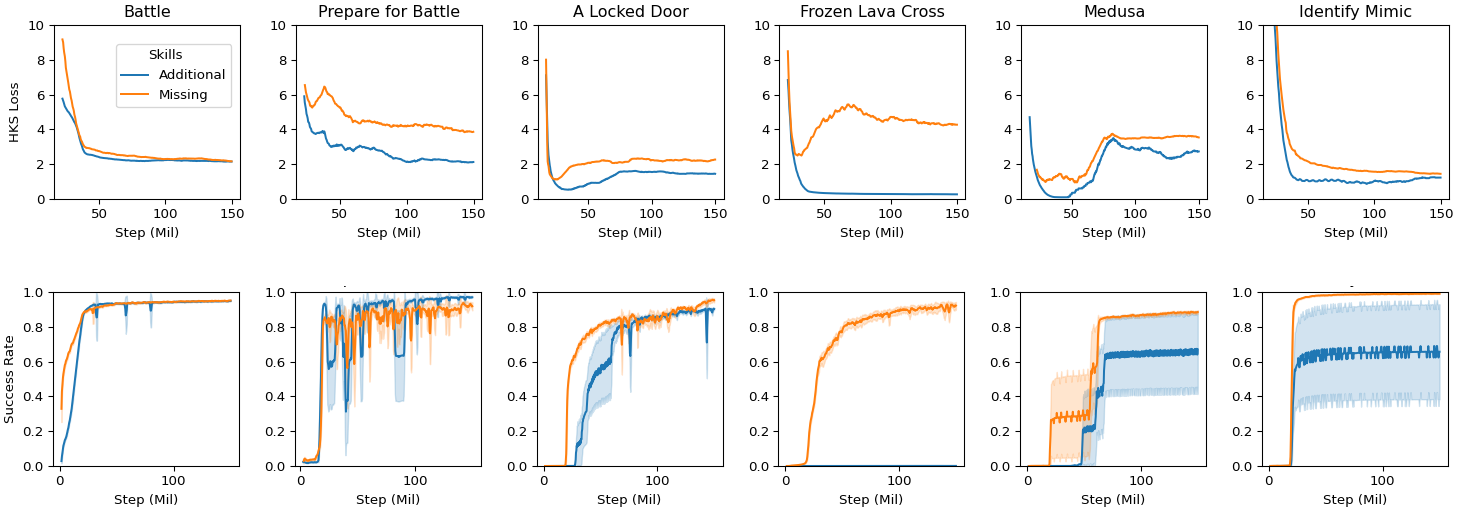}
    \caption{The first row shows HKS loss for the 6 SkillHack environments that were run with both missing and additional skills.  The second row shows success rate on these same tasks for reference.}
    \label{fig:hks_loss}
\end{figure}

Figure \ref{fig:hks_loss} shows the HKS loss over time for the experiments that were run with missing and additional skills.  Notably, the HKS loss tends to converge to a non-zero value, indicating that the agent policy would not perfectly match that of the policy-over-teachers.  This implies that the skills policies do not completely correspond to the policies needed in the task, which was to be expected, as the skill acquisition environments did not perfectly match the task environments.

The HKS loss for the experiments with additional skills tend to converge to a smaller value than those in which a skill was removed.  This is because the greater range of skills provided in these experiments allows the policy-over-teacher to more accurately match the agent policy.  In the cases where skills have been removed, the policy-over-teachers is again likely substituting in the closest matching skill during the timesteps taken up by the missing skill.

A notable result is seen on the experiment for \task{A Locked Door} with additional skills.  As can be seen from the corresponding success rate plot, this experiment failed to converge.  The HKS loss converges at close to 0, implying that the agent policy has almost perfectly synced with the policy-over-teachers.  In the absence of external rewards, the agent policy will only be shaped by the intrinsic losses and, ignoring other intrinsic losses (the IMPALA agent does include some other intrinsic losses but these are roughly a magnitude smaller than the HKS loss), this means that the agent policy will be almost wholly shaped by the HKS loss.  Therefore it logically follows that it would learn to minimise this loss and perfectly match the policy-over-teachers, which is what we see in the plot.  What this plot does not show is whether the skills being chosen by the policy-over-teachers is changing during training on this experiment or whether they are remaining constant.  It seems likely that the case is the latter, as if the agent was changing the skills it used over time, it would likely eventually reach the correct set of skills and therefore bias the exploration of the agent towards succeeding on the task.  This could be an indication that in the case when the initial exploratory period fails to find any reward, the agent and policy-over-teachers get stuck performing the same set of skills.

\subsection{Limitations}

Despite the relative success of the skill based methods over the vanilla RL baseline, each one fails to learn on at least two of the SkillHack tasks.  As discussed previously, a significant reason for this may be that badly designed skills do not transfer to the desired task, providing no useful prior for exploration.  Indeed, a badly designed skill will likely actually harm exploration, by inducing the agent to explore irrelevant action sequences.  This is all to say that these methods are only as good as the skills that are provided to them, although they have shown some ability to be robust to missing and useless skills, provided that the rest of the skill set is well crafted.  The reliance on predefined skills can be seen as a fundamental limitation of the skill based methods, as they require domain knowledge of the task to engineer useful skills or skill acquisition environments.  However, if this domain knowledge is present, we believe that skill based methods provide a practical way to inject this information into an agent.  
\section{Related Work}

\paragraph{Policy Transfer}  
Policy transfer is defined as the paradigm where the learning process of an agent is supplemented by utilizing policies pretrained on related tasks.  It is a paradigm that becomes more prevalent as environments develop larger state-action spaces, sparser rewards and more open-ended sets of possible behaviours.
\textit{Knowledge distillation} is a popular technique in supervised learning that ensembles knowledge from many teacher models into a student model \citep{Hinton2015DistillingTK}.
Recent work in RL has used the distillation of policies to transfer knowledge from teacher policies to the student. \citet{RusuCGDKPMKH15} propose \textit{policy distillation} where the student policy is trained by minimising the divergence between the action distributions of the teacher and student policies over the trajectories sampled via the teacher policies.
\citet{czarnecki19a} performs policy distillation using the trajectories that are sampled with the student policy, rather than teachers'. 
Another objective for policy transfer is minimising the cross-entropy between the teacher and student policies, as is done in Kickstarting RL \citep{schmitt2018kickstarting}.
The Actor-Mimic algorithm \citep{ParisottoBS15} takes a slightly different approach to Kickstarting.  First, the student is pretrained to master all the skill environments, but with the coefficient governing this loss function held constant.  The resultant weights are used to initialise a network that then trains on the target environment, without any supervision from the teachers.
A different perspective for achieving transfer is maximising the probability that the student and teacher policies will visit the same trajectories. 
An example of such an approach is \textit{Distral} \citep{teh2017distral} where a central ``distilled" policy is shared across many tasks in order to capture common behaviour. 
Each teacher is trained in separation to solve its own task while being constrained to stay close to the shared policy.

\paragraph{Hierarchical Reinforcement Learning}  
While we choose to use the Options Framework as a baseline and as a building block of HKS, many other HRL approaches exist.  Feudal RL \citep{dayan1992feudal} formulates an arbitrarily deep hierarchy of \textit{managers}.  The control hierarchy is such that each manager has a super-manager that sets it tasks and a set of sub-managers that it delegates tasks to.  A recent successful realisation of this framework is FeUdal Networks \citep{vezhnevets2017feudal}, in which a manager sets goals in a learned latent space.  Hierarchical Deep Q-Networks \citep{kulkarni2016hierarchical} proposes a hierarchical structure for value based RL. A meta-controller produces a value function over goals conditioned on the state, while a controller produces a value function over actions, conditioned on the state and the chosen goal. A critic checks if the goal is reached and provides intrinsic rewards to the controller, which terminates either at the end of the episode or when the goal is reached, in which case the meta-controller picks a new goal.  
HIRO \citep{nachum2018dataefficient} is another 2 level hierarchical model, but with a focus on off-policy learning. Off-policy learning in HRL is especially complex, as changes to lower level policies correspond to changing the action space of higher level policies. The key insight to correct for this was to relabel old goals. Consider a buffer of states, goals, rewards and actions. To perform off-policy learning on this sequence, the goal is adjusted such that if the model was run with the current sub-policies, the same action distribution would be induced as in the memory buffer. A similar model is Hierarchical Actor-Critic \citep{levy2019LearningMH}, which additionally makes use of Hindsight Experience Replay to outperform HIRO.


\paragraph{Benchmarks} 
We consider a range of RL benchmarks and their viability to the problem of skill transfer.  Classic environments such as those in OpenAI gym \citep{brockman2016openai} are too simple to be broken down into constituent skills.  Common video game environments such as Atari \citep{bellemare2013atari}, DeepMind Lab \citep{beattie2016deepmind} and ViZDoom \citep{kempka2016vizdoom} are complex enough for skill-based learning to be warranted, but much of the focus of learning in these environments must be put towards interpreting the high dimensional pixel input.
The CORA benchmark \citep{powers2021cora} for continual learning proposes schedules of environments which are each shown to the agent for some large number of iterations, with one of the primary goals being to \textit{forward transfer} knowledge from previously seen environments to new ones.  Skill transfer could be seen as a subset of continual learning, where the focus is entirely on the forward transfer of information from skills to tasks.


The MineRL \citep{guss2021minerl} benchmark challenges the RL agent to mine a diamond in the popular video game Minecraft. Using vanilla RL to do this would be intractable, so the benchmark comes with a dataset of over 60 million state-action tuples of recorded human demonstrations.  It should be noted that while similar to the use of expert teachers in SkillHack, new samples cannot be dynamically generated from the teachers, meaning this dataset could not be used by any of the methods we investigate.
%
Crafter \citep{hafner2022benchmarking} is a 2D open world survival game, with similar game dynamics to Minecraft but with a lower dimensional input space and faster runtime.  
The paper formulates a taxonomy of 22 achievements and their dependencies on each other.  
By interpreting achievements with dependencies as tasks and achievements with no dependencies as skills, Crafter could be phrased as a skill transfer benchmark.  
However, the majority of the tasks included only have 1 associated skill and many skills have no associated task.




\vspace{-3mm}
\section{Conclusion}
\vspace{-2mm}

This paper presents SkillHack: a NetHack-based benchmark for skill transfer in RL. SkillHack features 16 skill environments for acquiring knowledge necessary to solve 8 target problems, all of which have been shown to be intractable for vanilla RL.  Our experimental results show that the SkillHack tasks form a difficult and varied suite of challenges, with methods performing markedly differently across the range of tasks.
We also propose \methodfull{} (\method{}): a new method for skill transfer that combines Kickstarting and the Options Framework.
Evaluating \method{} on SkillHack shows that it improves the final performance on the overall benchmark by $15\%$ over the next best baseline (Kickstarting).

In the near future, we aim to conduct additional experiments to compare the methods by conditioning on experts for all 16 skill environments, rather than the subset of relevant experts for each task.  
We would also like to investigate extending our baseline for the Options Framework to allow for options to learn during training \citep{bacon2016option}.
Finally, we plan to use SkillHack to study other policy transfer approaches including inter-task mapping, representation transfer and transfer learning approaches for the framework of unsupervised RL.

We hope that SkillHack can serve as a useful benchmark in the community for evaluating skill transfer and that \method{} can find applications where traditional RL methods struggle.  We are open sourcing both the SkillHack benchmark and the \method{} method and look forward to contributions from the community.
\section*{Acknowledgements}

We would like to thank Yicheng Luo, Minqi Jiang, and Sam Powers for many valuable discussions during the course of completing this work.  Furthermore, we would like to thank our anonymous reviewers for their insightful feedback and recommendations for improving the paper.

\bibliography{collas2022_conference}
\bibliographystyle{collas2022_conference}

\newpage
\appendix
\section{Action Space} \label{app:nn}

NetHack takes ASCII characters as its inputs.  These can be modified with certain keys like \keystroke{Ctrl} and \keystroke{Shift}.  We use a reduced action space of size 32, only allowing actions that have been deemed relevant for completing the benchmark.  This includes 16 movement actions consisting of the 4 cardinal directions, the 4 diagonal directions each with a single and long movement option.  The remaining actions are summarised in Table \ref{tab:actions}.
\begin{table*}[t]
    \centering
    \begin{tabular}{@{}l c@{}} 
     \toprule
     \textbf{Name} & \textbf{Key} \\
     \midrule
     PickUp & \keystroke{,} \\
     PutOn & \keystroke{Shift} + \keystroke{p} \\
     Zap & \keystroke{z} \\
     TakeOff & \keystroke{Shift} + \keystroke{t} \\
     Wear & \keystroke{Shift} + \keystroke{w} \\
     Throw & \keystroke{t} \\
     Escape & \keystroke{Esc} \\
     Eat & \keystroke{e} \\
     Apply & \keystroke{a} \\
     Wield & \keystroke{w} \\
     Quaff & \keystroke{q} \\
     Gold & \keystroke{\$} \\
     Inv1 & \keystroke{f} \\
     Inv2 & \keystroke{g} \\
     Inv3 & \keystroke{h} \\
     \bottomrule
	\end{tabular}
	\caption{Command Actions}
    \label{tab:actions}
\end{table*}

\section{Kickstarting Coefficient Discussion} \label{app:ks_sched}

The Kickstarting coefficient $\lambda$ was set with a manual schedule defined in Section \ref{sec:baselines}.  Notably, we never let the coefficient drop to 0, but instead specify a minimum value of $0.1$, meaning that the Kickstarting loss is always present.  The reason for this was twofold.  Firstly, we saw that large spikes in performance would often occur after the coefficient had reached its minimum value of $0.1$.  Therefore, keeping a small minimum allowed Kickstarting and HKS to work without going through the lengthy process of tuning the schedule for each environment.  Secondly, when the coefficient was allowed to drop to 0, we would see infrequent massive drops in performance with both Kickstarting and HKS (Around 1 in 2 runs would at some point lose around half of their performance after converging).  The reason for this is not entirely clear, but is an area of current investigation.

This artificial minimum on the coefficient effectively imposes a cap on the performance of Kickstarting, as it is always slightly being optimised away from following extrinsic reward.  HKS is not as susceptible to this effect, as the policy-over-teachers can converge such that the weighted average of teacher action distributions closely matches that which would maximise extrinsic reward anyway.  This can be seen in the plot of model performances (Figure \ref{fig:results}) for the tasks \task{Battle}, \task{Prepare for Battle} and \task{Identify Mimic}.  In all of these tasks, HKS converges to a slightly higher success rate than Kickstarting.  To account for this discrepancy, we take the average percentage difference in the final performances of Kickstarting and HKS on these three tasks, which works out to be $2.2\%$.  The final performance for Kickstarting averaged over all tasks is $39.9\%$.  We therefore increase this value by $2.2\%$ giving us $40.8\%$, giving us a fairer comparison against HKS. 

\section{Skill Listing} \label{app:skill_listing}
\begin{figure}[ht!]
\centering

\begin{subfigure}{.5\textwidth}
    \centering
    \includegraphics[width=0.9\textwidth]{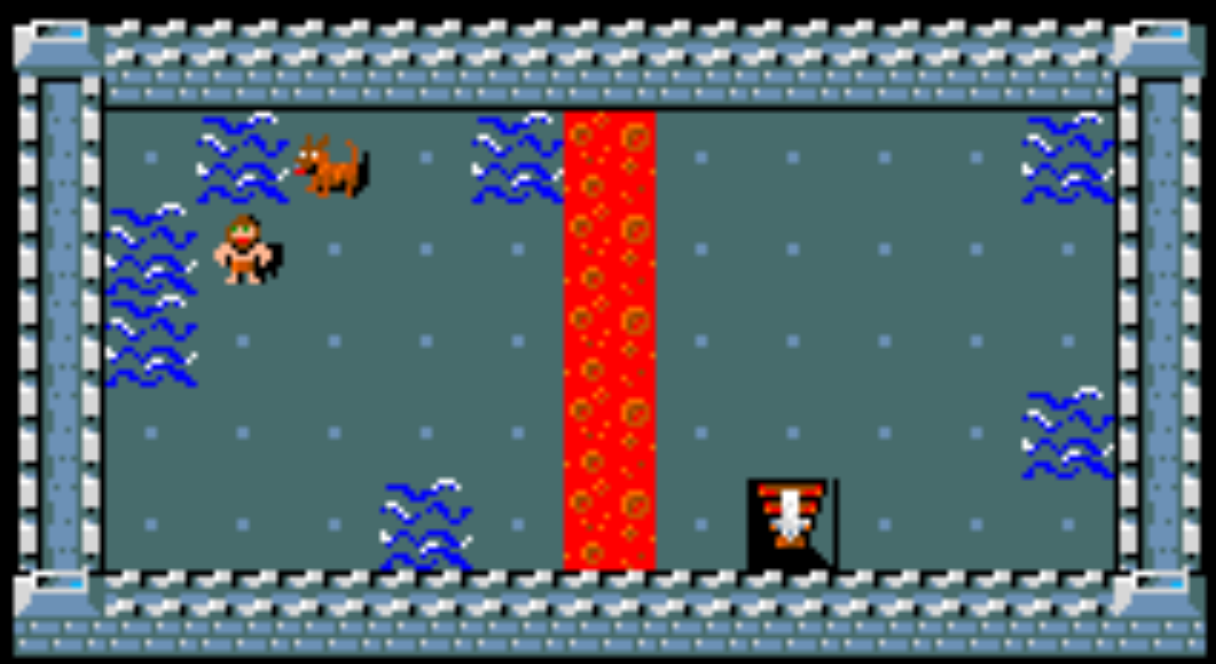}
    \caption{\skill{ApplyFrostHorn}}
\end{subfigure}%
\begin{subfigure}{.5\textwidth}
    \centering
    \includegraphics[width=0.9\textwidth]{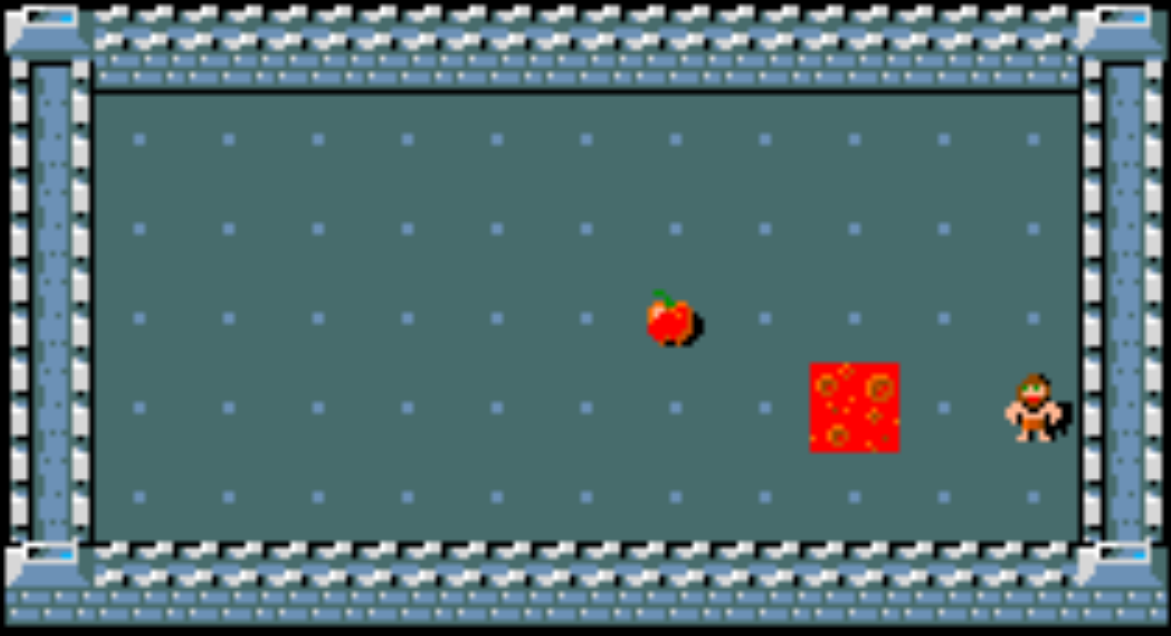}
    \caption{\skill{Eat}}
\end{subfigure}

\begin{subfigure}{.5\textwidth}
    \centering
    \includegraphics[width=0.9\textwidth]{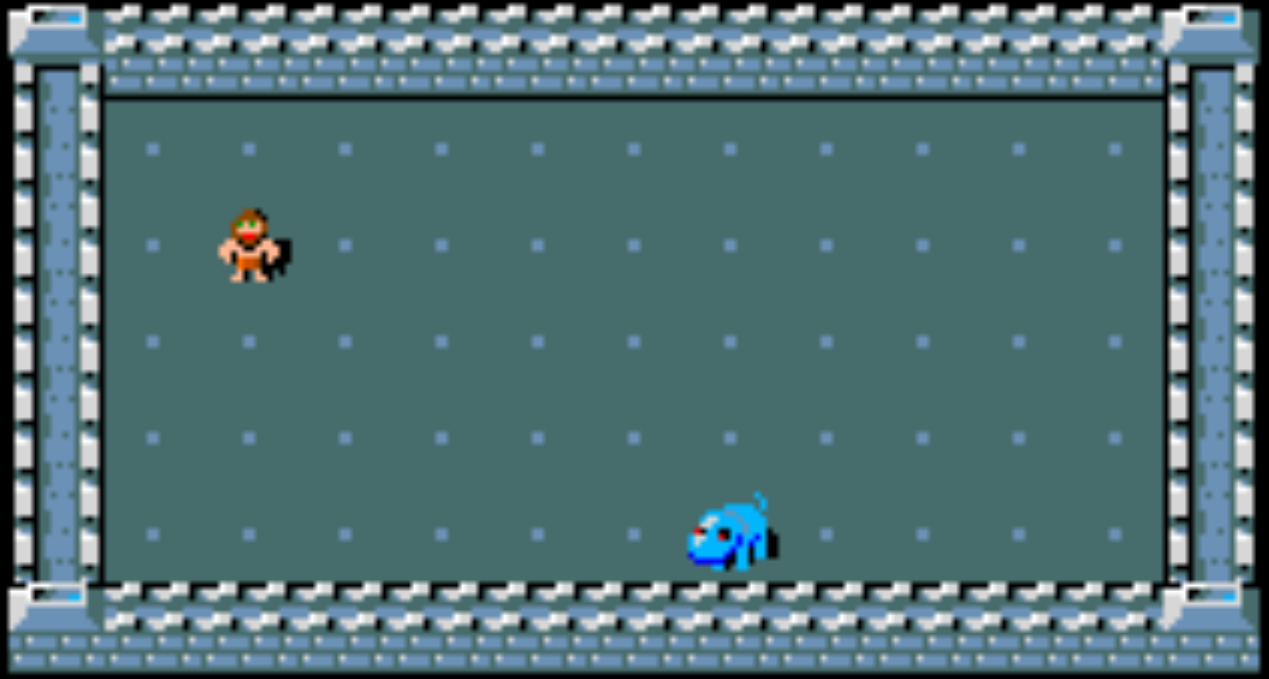}
    \caption{\skill{Fight}}
\end{subfigure}%
\begin{subfigure}{.5\textwidth}
    \centering
    \includegraphics[width=0.9\textwidth]{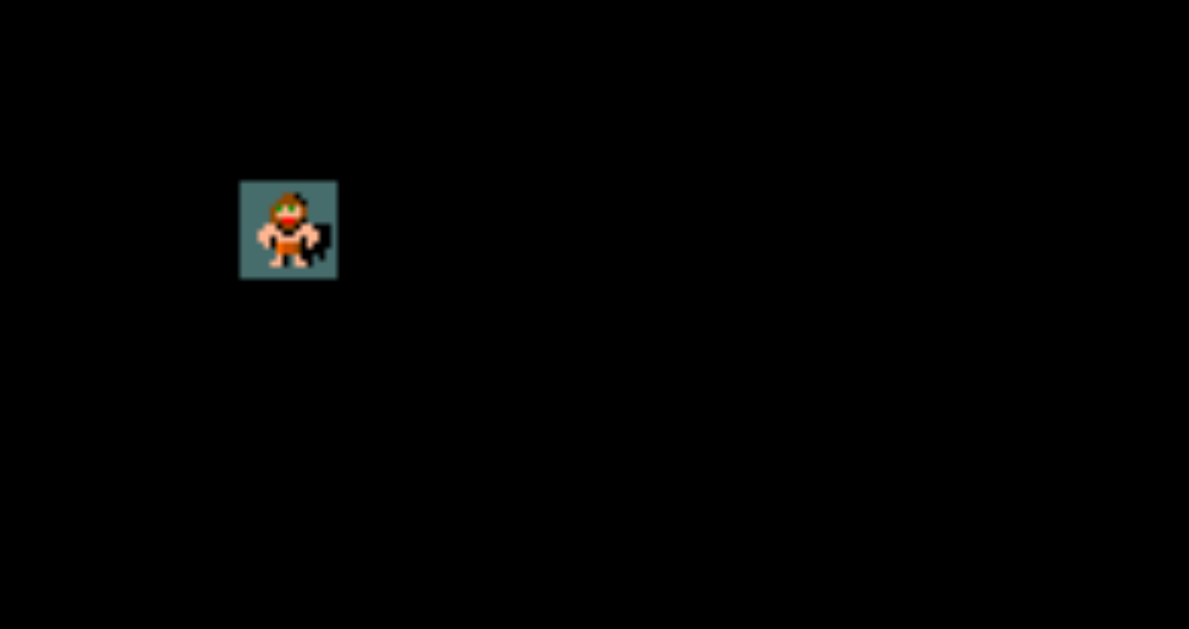}
    \caption{\skill{NavigateBlind}}
\end{subfigure}

\begin{subfigure}{.5\textwidth}
    \centering
    \includegraphics[width=0.9\textwidth]{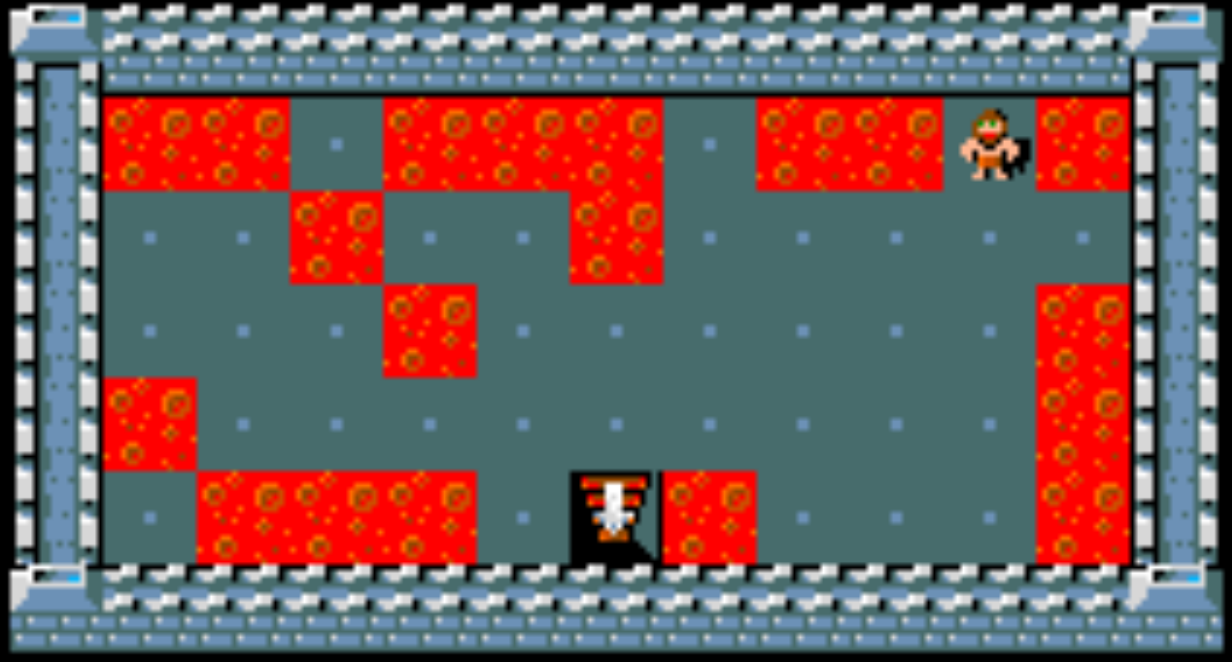}
    \caption{\skill{NavigateLava}}
\end{subfigure}%
\begin{subfigure}{.5\textwidth}
    \centering
    \includegraphics[width=0.9\textwidth]{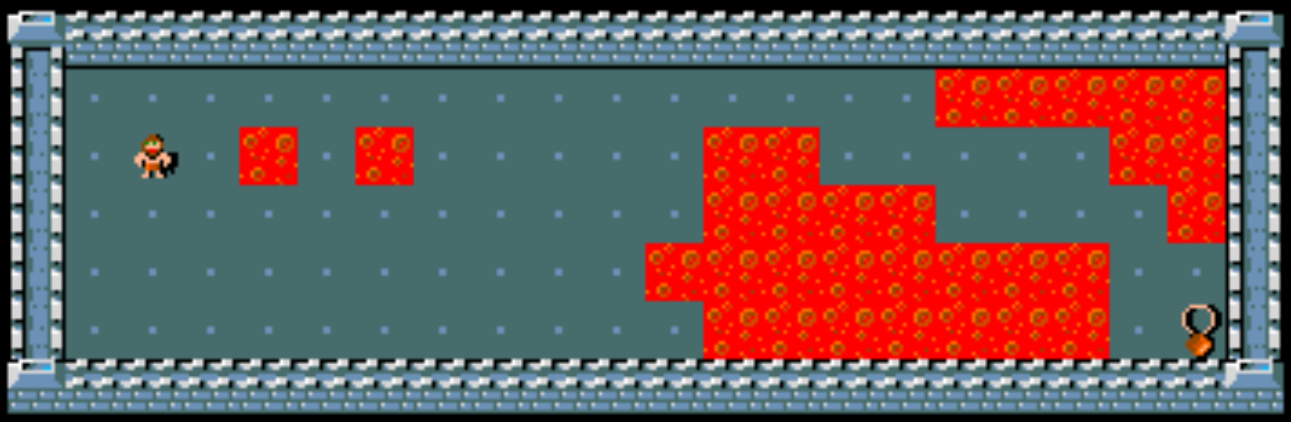}
    \caption{\skill{NavigateLavaToAmulet}}
\end{subfigure}

\begin{subfigure}{.5\textwidth}
    \centering
    \includegraphics[width=0.9\textwidth]{res/skills/nav_water.png}
    \caption{\skill{NavigateWater}}
\end{subfigure}%
\begin{subfigure}{.5\textwidth}
    \centering
    \includegraphics[width=0.9\textwidth]{res/skills/pick_up.png}
    \caption{\skill{PickUp}}
\end{subfigure}

\caption[short]{Skill environments.}
\label{fig:skill_envs_visual}
\end{figure}

\begin{figure}[ht!]
\centering

\begin{subfigure}{.5\textwidth}
    \centering
    \includegraphics[width=0.9\textwidth]{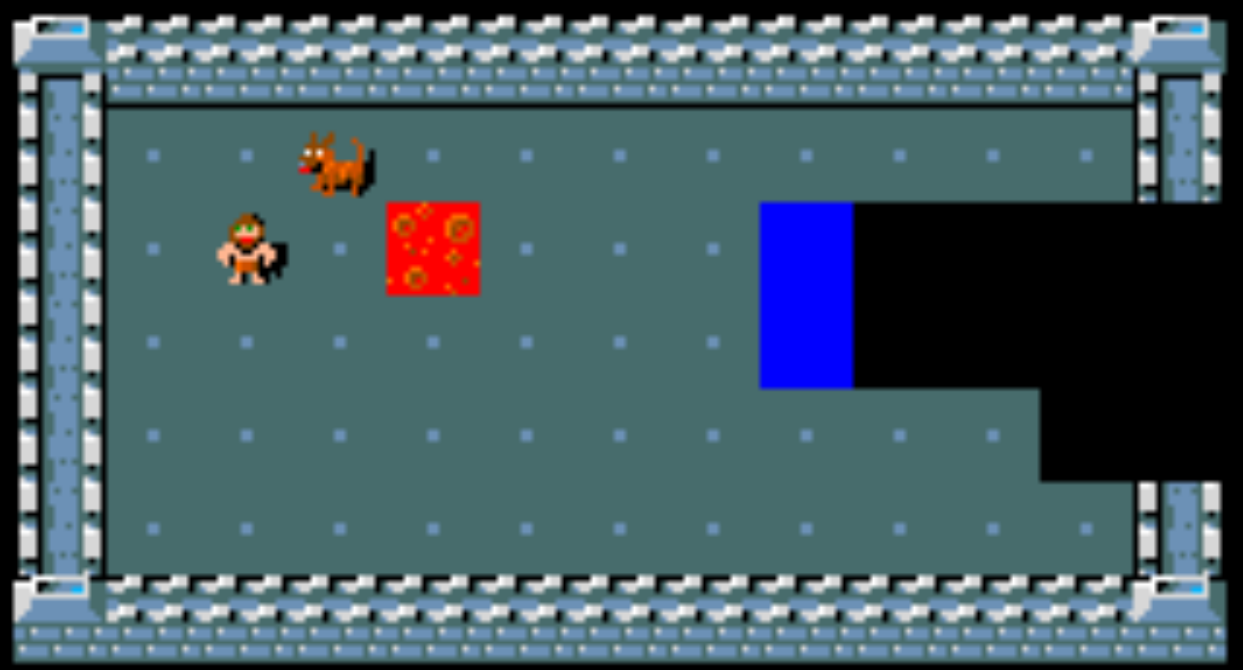}
    \caption{\skill{PutOn}}
\end{subfigure}%
\begin{subfigure}{.5\textwidth}
    \centering
    \includegraphics[width=0.9\textwidth]{res/skills/take_off.png}
    \caption{\skill{TakeOff}}
\end{subfigure}

\begin{subfigure}{.5\textwidth}
    \centering
    \includegraphics[width=0.9\textwidth]{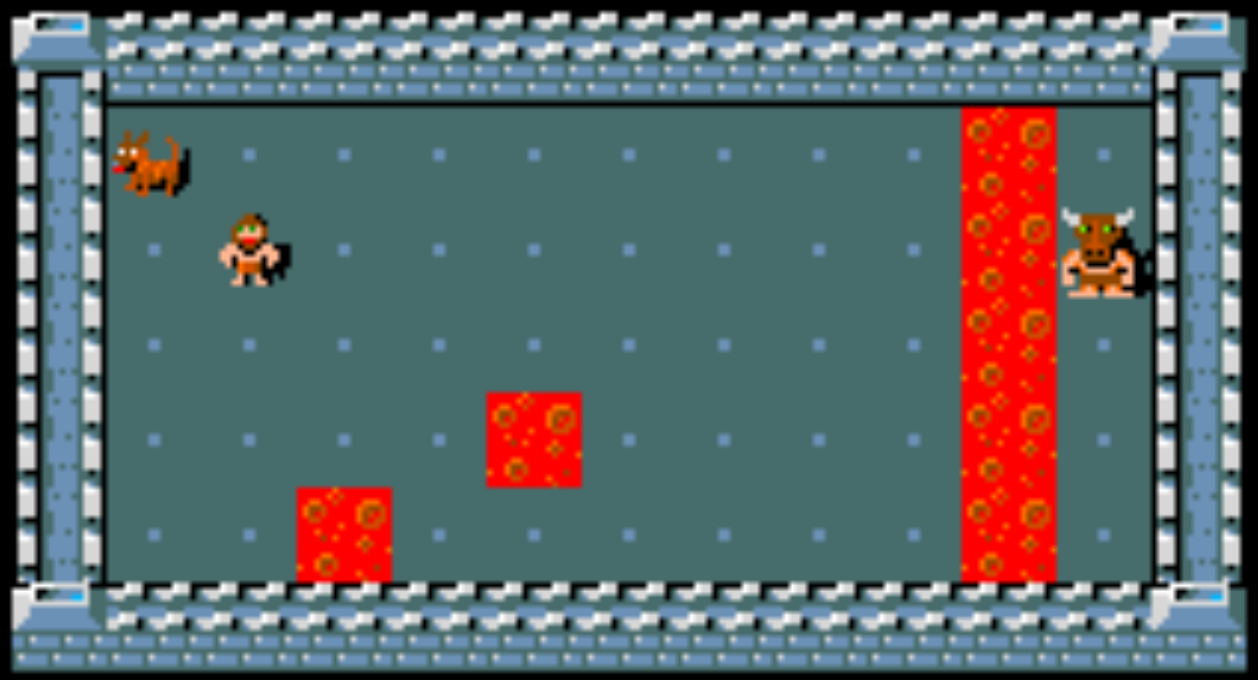}
    \caption{\skill{Throw}}
\end{subfigure}%
\begin{subfigure}{.5\textwidth}
    \centering
    \includegraphics[width=0.9\textwidth]{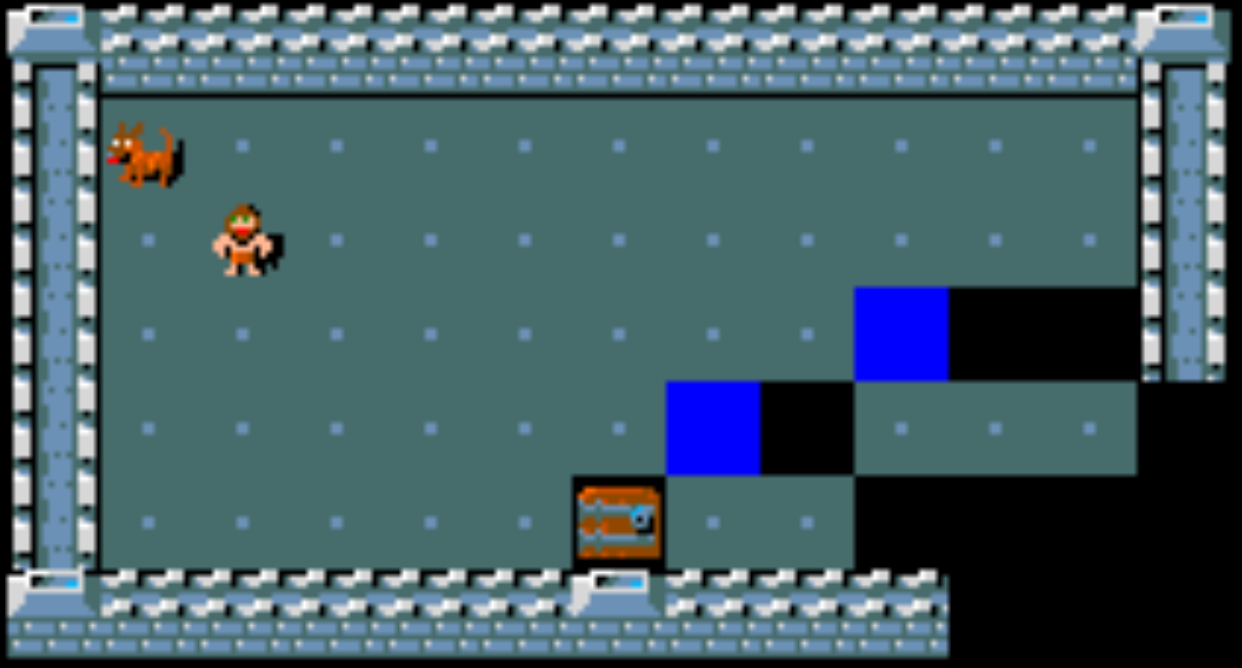}
    \caption{\skill{Unlock}}
\end{subfigure}

\begin{subfigure}{.5\textwidth}
    \centering
    \includegraphics[width=0.9\textwidth]{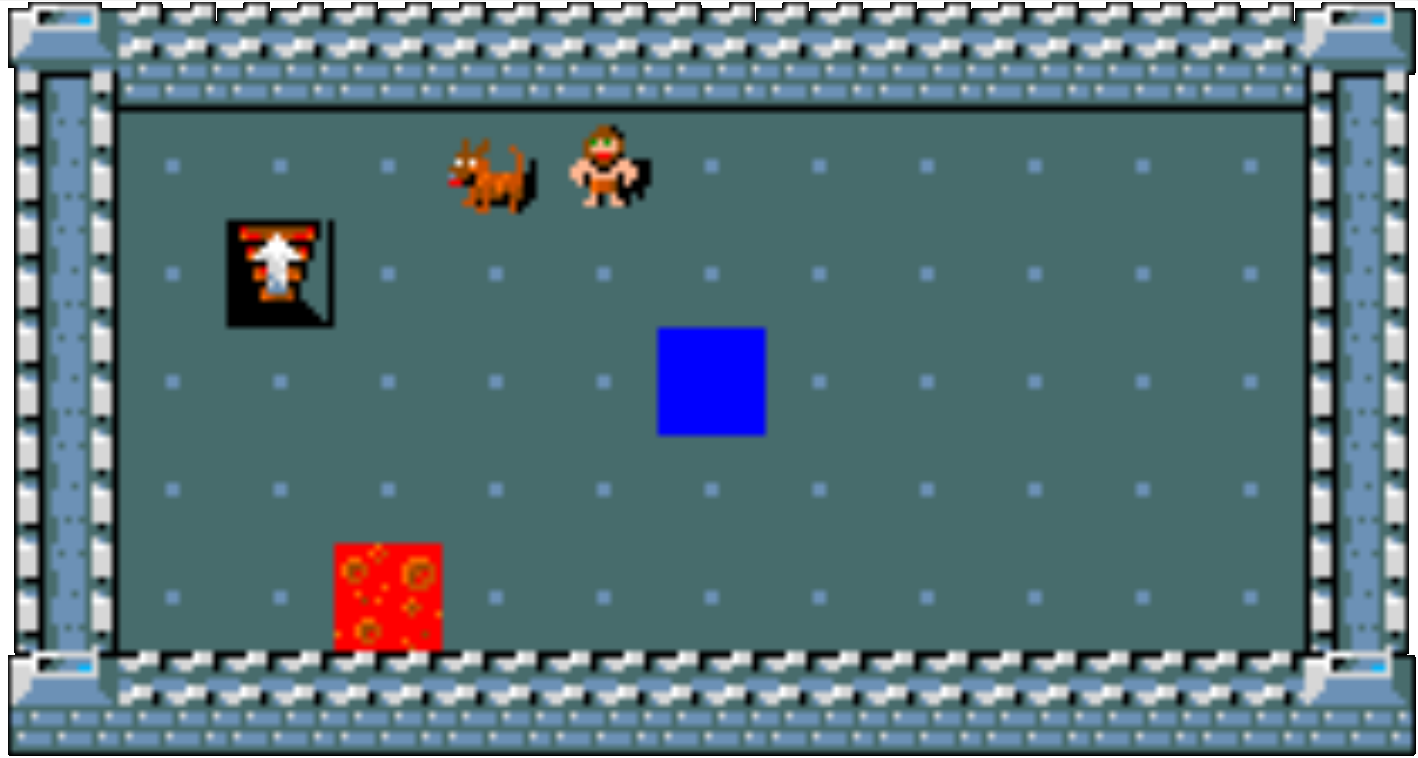}
    \caption{\skill{Wear}}
\end{subfigure}%
\begin{subfigure}{.5\textwidth}
    \centering
    \includegraphics[width=0.9\textwidth]{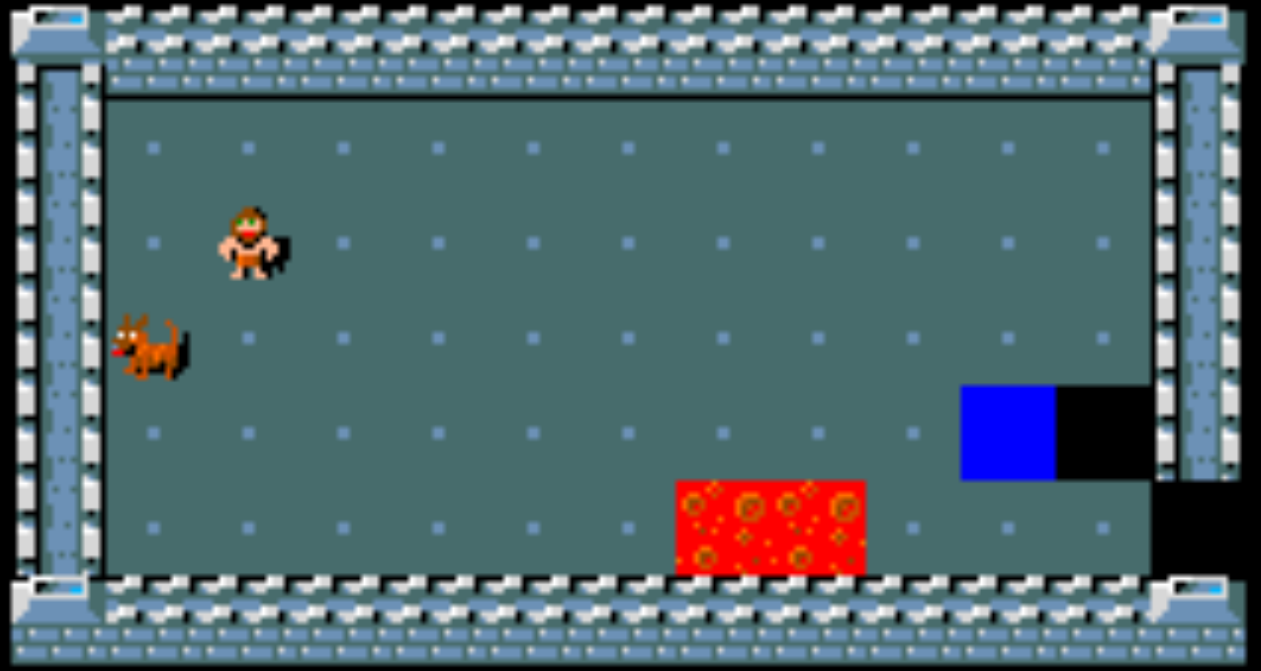}
    \caption{\skill{Wield}}
\end{subfigure}

\begin{subfigure}{.5\textwidth}
    \centering
    \includegraphics[width=0.9\textwidth]{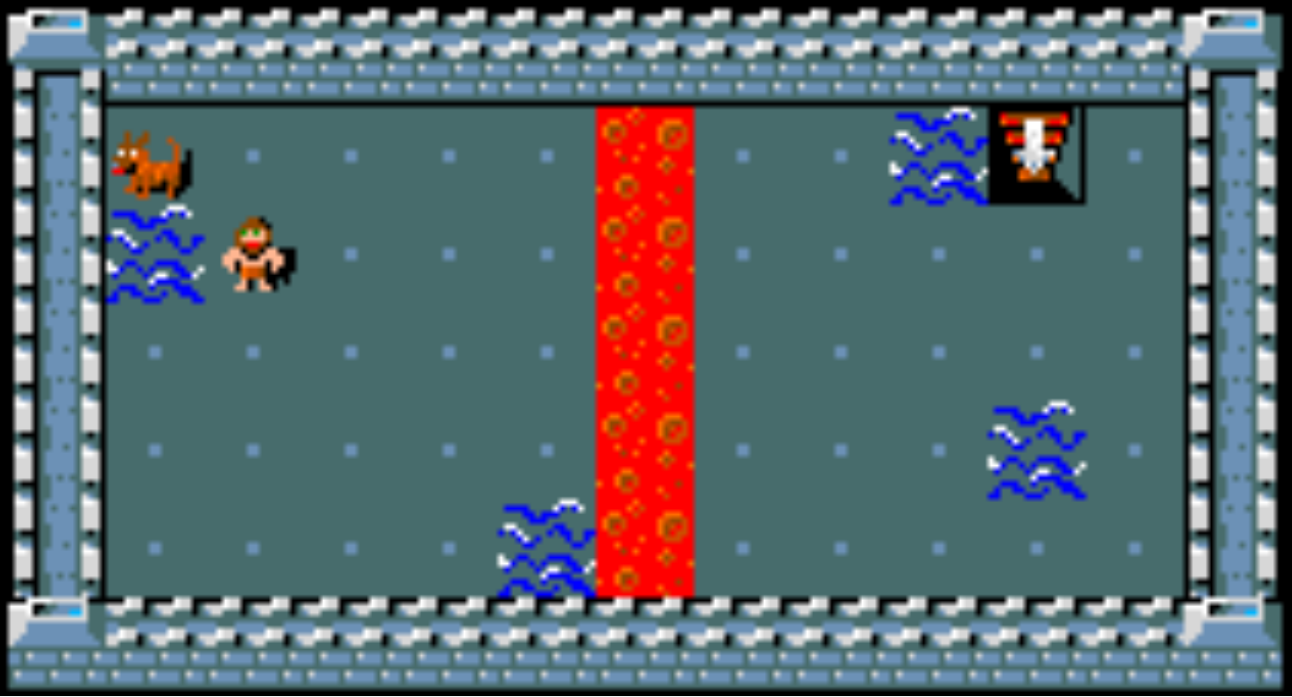}
    \caption{\skill{ZapWandOfCold}}
\end{subfigure}%
\begin{subfigure}{.5\textwidth}
    \centering
    \includegraphics[width=0.9\textwidth]{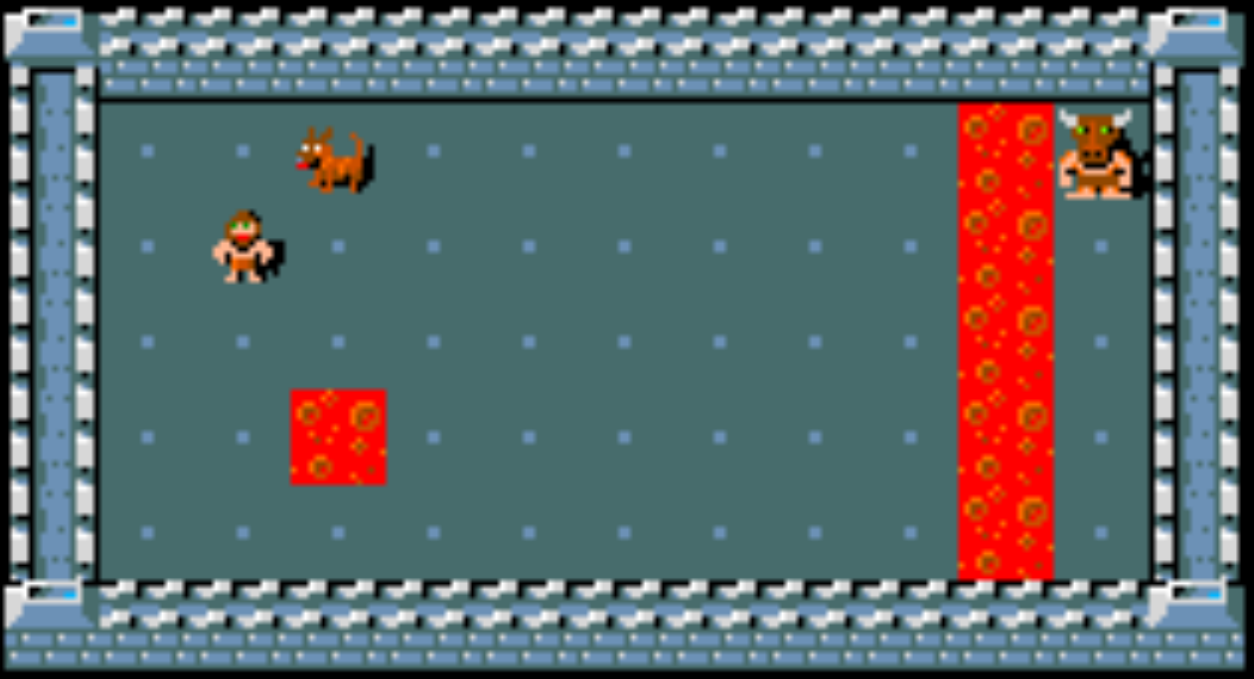}
    \caption{\skill{ZapWandOfDeath}}
\end{subfigure}

\caption[short]{Skill environments (continued).}
\label{fig:skill_envs_visual2}
\end{figure}

The list of skills is shown in Table \ref{tab:skills}.  Each skill has an environment where it is learned in isolation, shown in Figures \ref{fig:skill_envs_visual} and \ref{fig:skill_envs_visual2}.  It should be noted that many of the environments visually look the same.  This is because many environments focus on manipulating items in the players inventory, with the rest of the world only serving as a distraction which the agent must learn to reliably ignore.
\begin{table*}[t]
    \centering
    \begin{tabular}{@{}l l@{}} 
     \toprule
     \textbf{Name} & \textbf{Description} \\
     \midrule
     \skill{ApplyFrostHorn} & Use a frost horn to freeze some lava. \\
     \skill{Eat} & Eat an apple. \\
     \skill{Fight} & Hit a monster. \\
     \skill{NavigateBlind} & Reach the staircase while blinded.\\
     \skill{NavigateLava} & Reach the staircase past random lava patches. \\
     \skill{NavigateLavaToAmulet} & Reach an amulet past random lava patches. \\
     \skill{NavigateWater} & Reach the staircase past random water patches. \\
     \skill{PickUp} & Pick up a random item. \\
     \skill{PutOn} & Put on an amulet or towel. \\
     \skill{TakeOff} & Take off clothes. \\
     \skill{Throw} & Throw daggers at a statue or at a monster. \\
     \skill{Unlock} & Use a key to unlock a locked door. \\
     \skill{Wear} & Wear a robe. \\
     \skill{Wield} & Wield a sword.\\
     \skill{ZapWandOfCold} & Use a wand of cold to freeze lava. \\
     \skill{ZapWandOfDeath} & Use a wand of death to kill a monster. \\
     \bottomrule
	\end{tabular}
	\caption{Skills contained in the benchmark.}
    \label{tab:skills}
\end{table*}

\section{Task Listing} \label{app:task_listing}

\begin{figure}[ht!]
\centering

\begin{subfigure}{0.45\textwidth}
    \centering
    \includegraphics[width=0.9\textwidth]{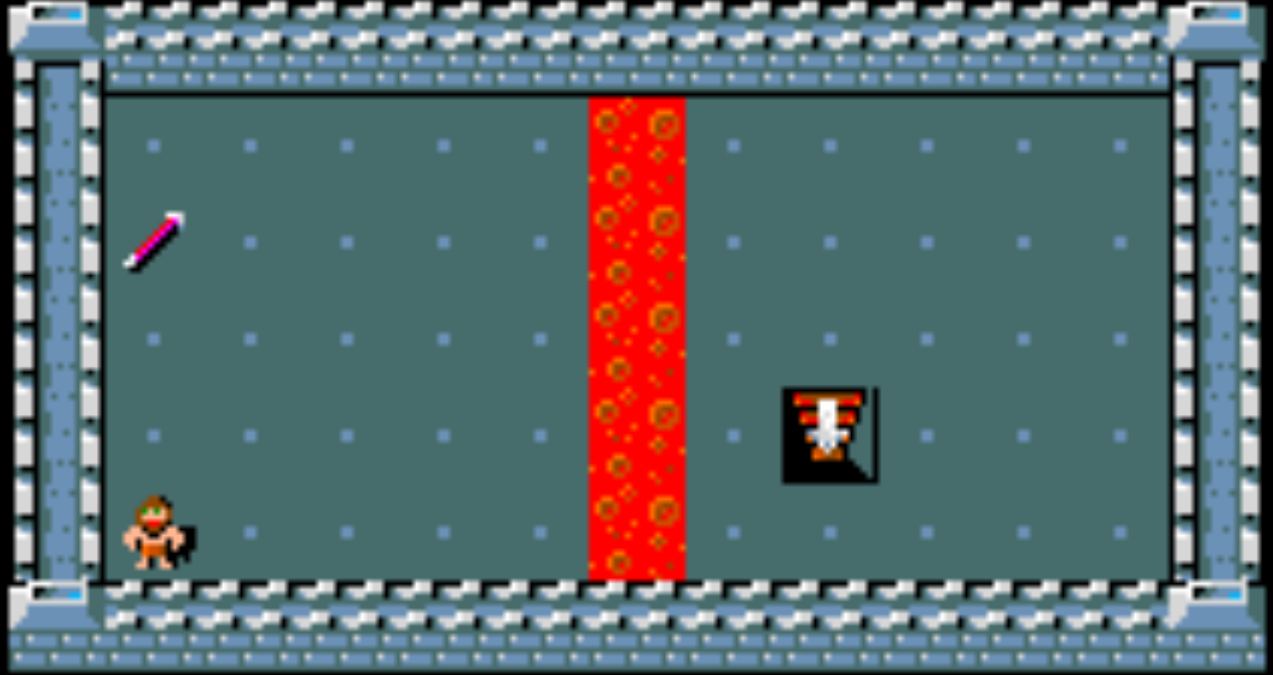}
    \caption{\task{Frozen Lava Cross}}
\end{subfigure}
\begin{subfigure}{0.45\textwidth}
    \centering
    \includegraphics[width=0.9\textwidth]{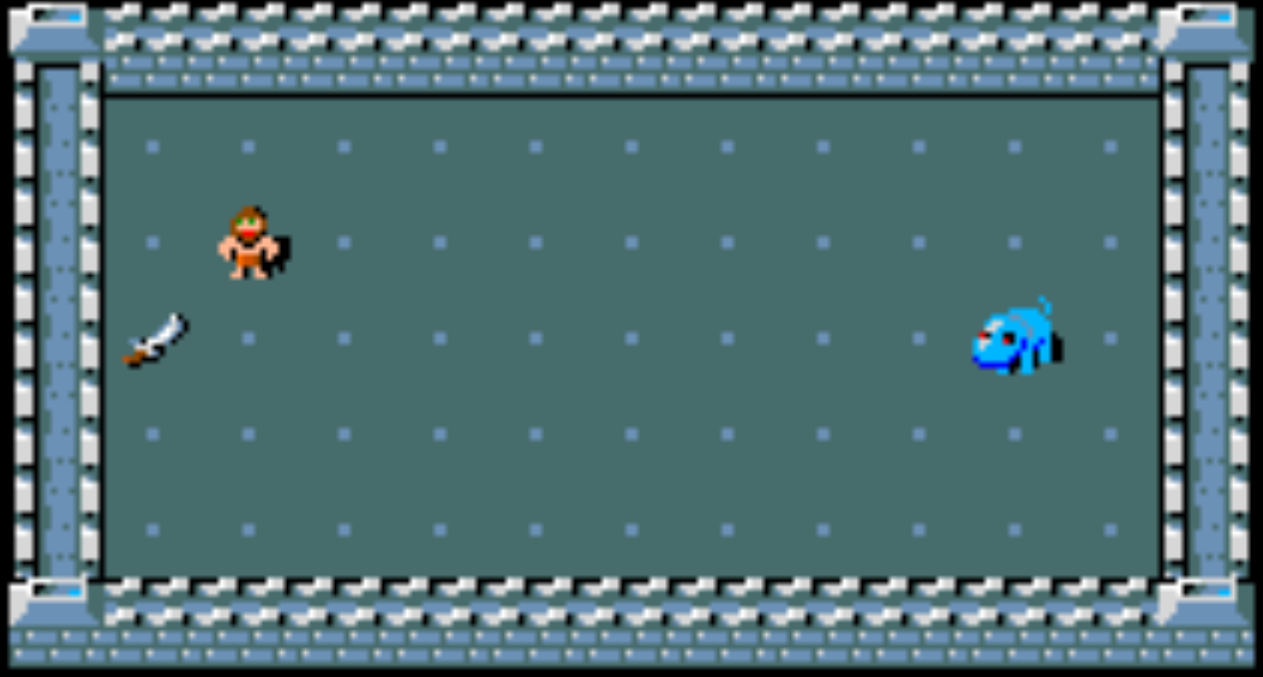}
    \caption{\task{Battle}}
\end{subfigure}

\begin{subfigure}{1\textwidth}
    \centering
    \includegraphics[width=0.9\textwidth]{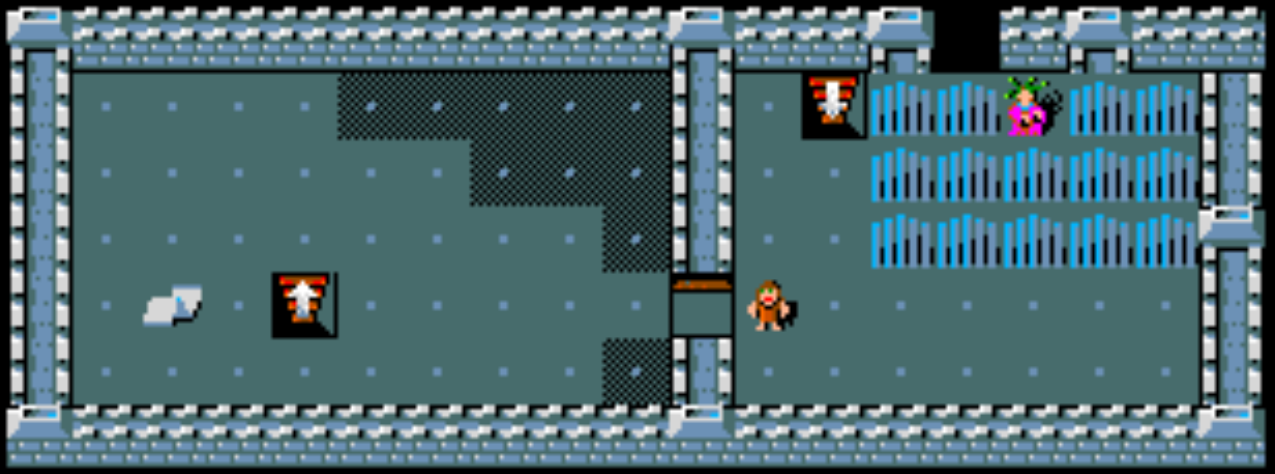}
    \caption{\task{Medusa}}
\end{subfigure}

\begin{subfigure}{1\textwidth}
    \centering
    \includegraphics[width=0.9\textwidth]{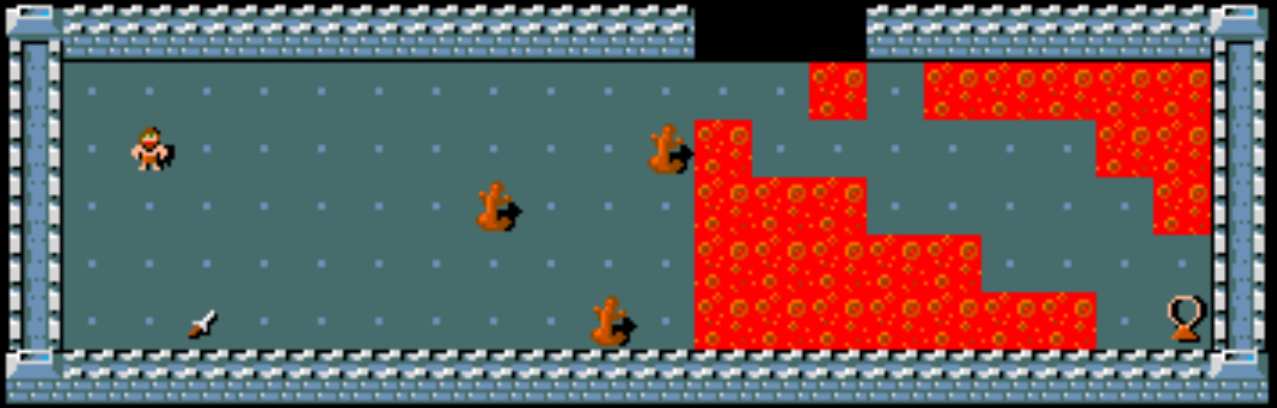}
    \caption{\task{Identify Mimic}}
    \label{fig:app_identify_mimic}
\end{subfigure}
\begin{subfigure}{1\textwidth}
    \centering
    \includegraphics[width=0.9\textwidth]{res/sea_monsters_cropped.png}
    \caption{\task{Sea Monsters}}
\end{subfigure}

\caption[short]{Task environments.}
\label{fig:tasks_visual}
\end{figure}

\begin{figure}[ht!]
\centering

\begin{subfigure}{0.45\textwidth}
    \centering
    \includegraphics[width=0.9\textwidth]{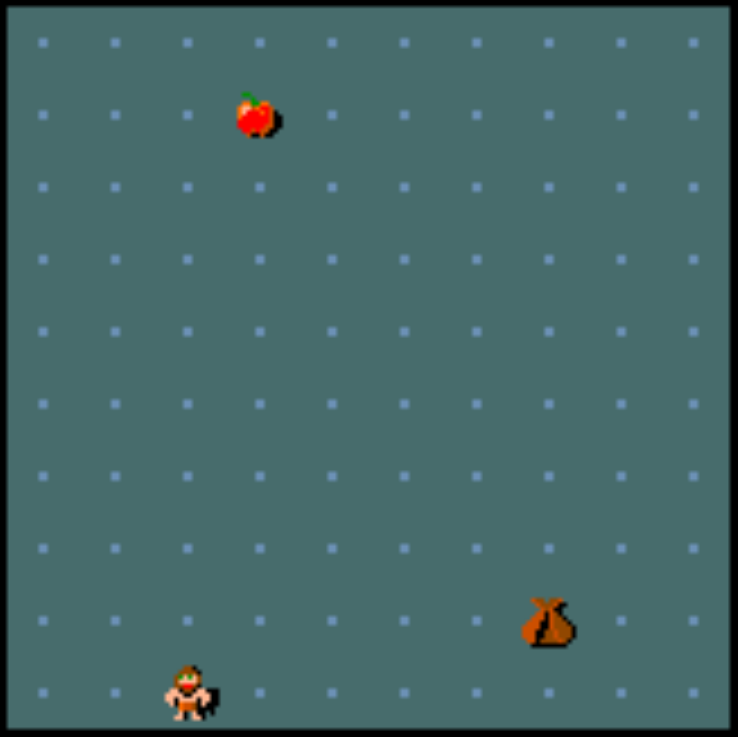}
    \caption{\task{Prepare For Battle}}
\end{subfigure}
\begin{subfigure}{0.45\textwidth}
    \centering
    \includegraphics[width=0.9\textwidth]{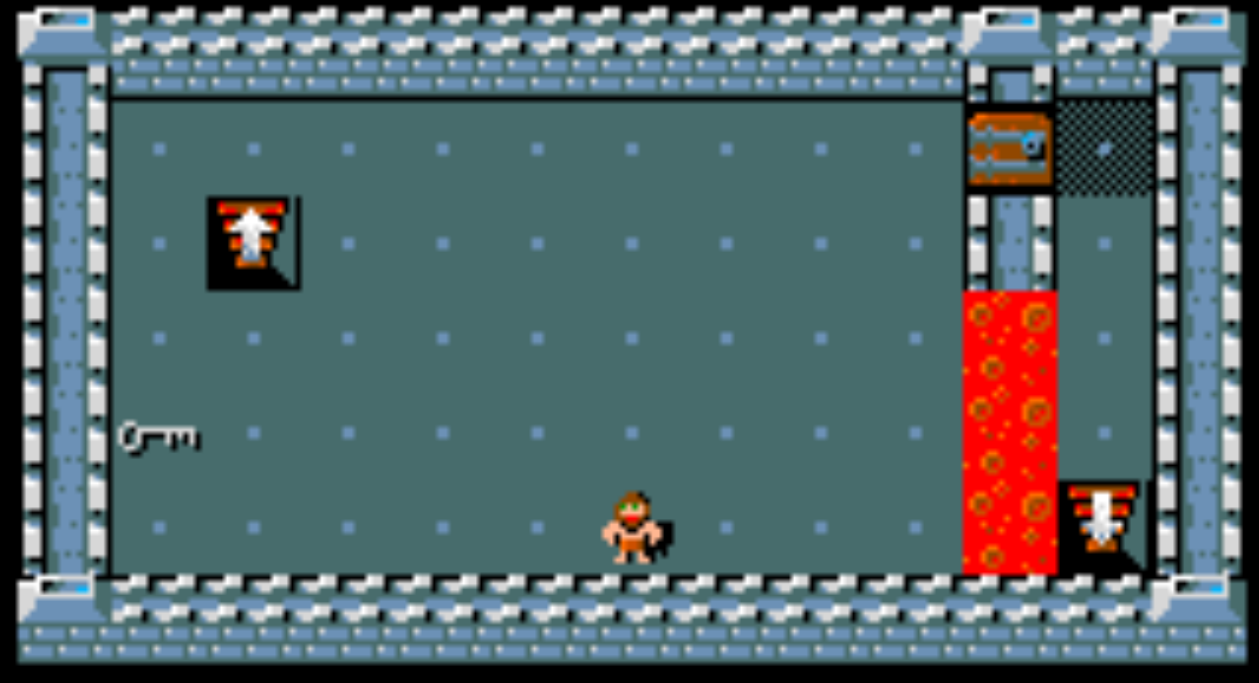}
    \caption{\task{A Locked Door}}
\end{subfigure}

\begin{subfigure}{1\textwidth}
    \centering
    \includegraphics[width=0.9\textwidth]{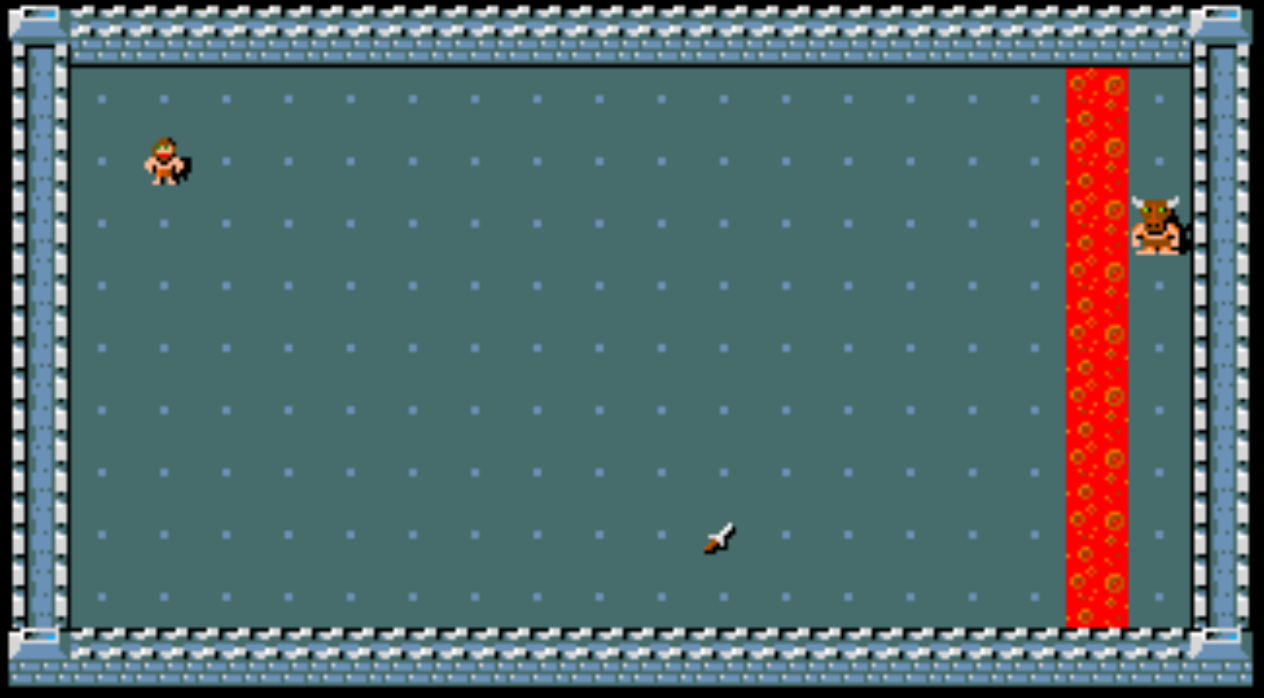}
    \caption{\task{Target Practice}}
\end{subfigure}

\caption[short]{Task environments (continued).}
\label{fig:tasks_visual2}
\end{figure}

The task descriptions can be found in Table \ref{tab:tasks}.  The environments are shown in Figures \ref{fig:tasks_visual} and \ref{fig:tasks_visual2}.

\section{Further Results} \label{app:further_results}

\begin{figure}
    \centering
    \includegraphics[width=1\textwidth]{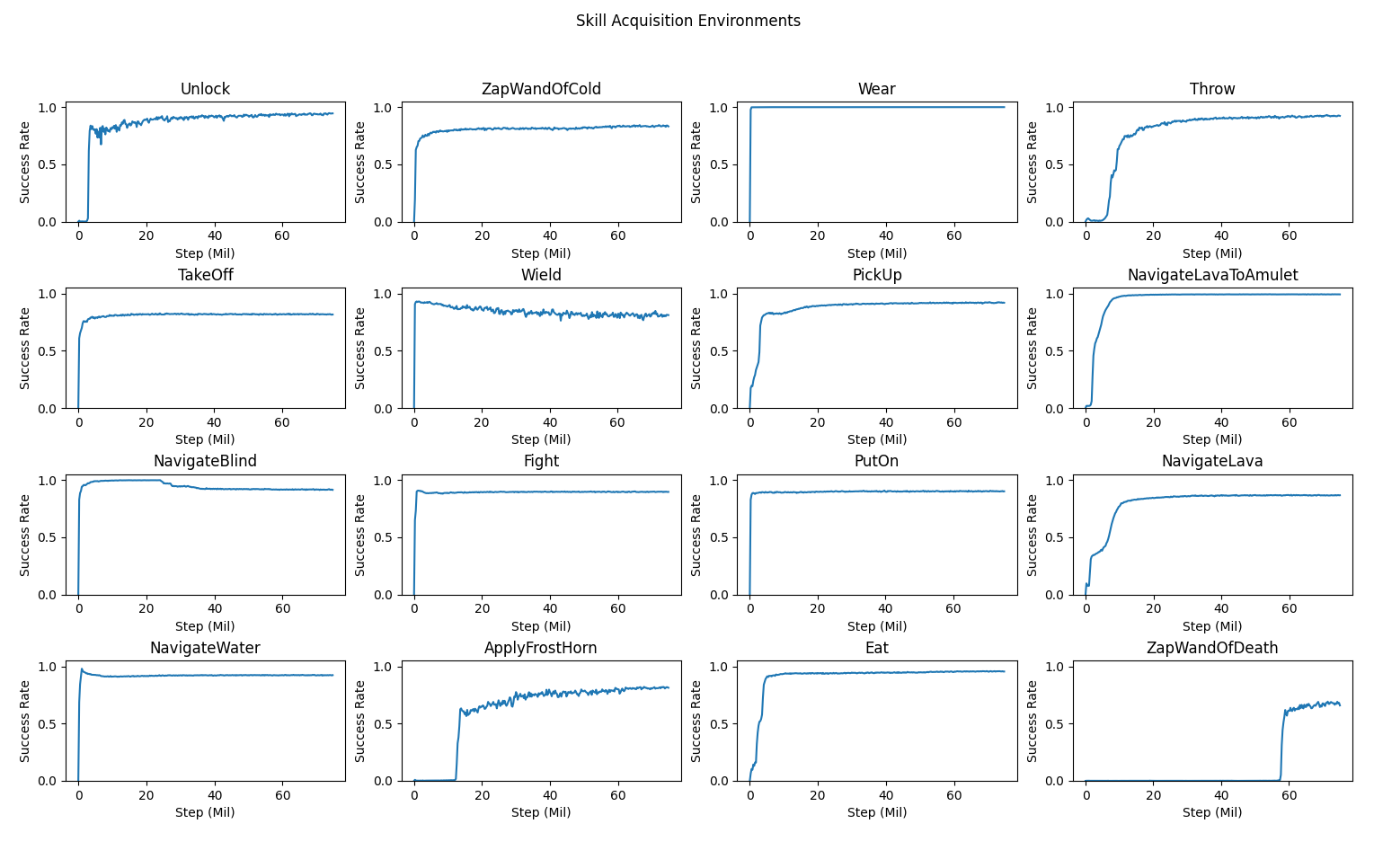}
    \caption{Success rate for training skill experts on the 16 skill acquisition environments}
    \label{fig:skill_graphs}
\end{figure}

The experts were all trained on the skill acquisition environments for $7.5 \times 10^7$ timesteps.  The results are shown in Figure \ref{fig:skill_graphs}.

\section{Hyperparameters} \label{app:hyp}
The hyperparameters for the IMPALA agent are summarised in Table \ref{tab:hyper}.
\begin{table*}[t]
    \centering
    \begin{tabular}{@{}l c@{}} 
     \toprule
     \textbf{Name} & \textbf{Value} \\
     \midrule
     \textbf{Training Settings} & \\
     \midrule
     num\_actors & 256 \\
     batch\_size & 32 \\
     unroll\_length & 80 \\
     \midrule
     \textbf{Model Settings} & \\
     \midrule
     hidden\_dim & 256 \\
     embedding\_dim & 64\\
     glyph\_type & all\_cat \\
     equalize\_input\_dim & false \\
     layers & 5 \\
     crop\_model & cnn \\
     crop\_dim & 9 \\
     use\_index\_select & true \\
     max\_learner\_queue\_size & 1024 \\
     \midrule
     \textbf{Loss Settings} & \\
     \midrule
     entropy\_cost & $0.001$ \\
     baseline\_cost & $0.5$ \\
     discounting & $0.999$ \\
     reward\_clipping & none \\
     normalize\_reward & true\\
     \bottomrule
	\end{tabular}
	\quad
	\begin{tabular}{@{}l c@{}} 
     \toprule
     \textbf{Name} & \textbf{Value} \\
     \midrule
     \textbf{Optimizer Settings} & \\
     \midrule
     learning\_rate & $0.0002$ \\
     grad\_norm\_clipping & 40\\
     \midrule
     \textbf{Intrinsic Reward Settings} & \\
     \midrule
     int$.$twoheaded & true \\
     int$.$input & full\\
     int$.$intrinsic\_weight & $0.1$ \\
     int$.$discounting & $0.99$ \\
     int$.$baseline\_cost & $0.5$ \\
     int$.$episodic & true \\
     int$.$reward\_clipping & none \\
     int$.$normalize\_reward & true \\
     \bottomrule
	\end{tabular}
	\caption{Hyperparameters of the IMPALA agent.}
    \label{tab:hyper}
\end{table*}

\end{document}